\documentclass{article} 
\usepackage{iclr2027_conference,times}

\usepackage{amsmath,amsfonts,bm}

\def\eqref#1{equation~\ref{#1}}

\def\1{\bm{1}}

\DeclareMathAlphabet{\mathsfit}{\encodingdefault}{\sfdefault}{m}{sl}
\SetMathAlphabet{\mathsfit}{bold}{\encodingdefault}{\sfdefault}{bx}{n}

\usepackage{hyperref}
\usepackage{url}
\usepackage{xcolor}
\usepackage{amssymb}
\newcommand{\imp}[1]{\textcolor{blue!50!black}{\textbf{#1}}}

\usepackage{booktabs}
\usepackage{graphicx}
\usepackage{colortbl}
\usepackage{enumitem}
\usepackage{wrapfig}
\usepackage{subcaption}
\usepackage{tabularx}
\usepackage{float}
\usepackage{etoc}

\title{
ECHO: Correcting Time Series Foundation Model Forecasts
with Historical Analogues and Errors
}

\author{
Boya Zhang$^{1,2}$,
Xilin Dai$^{3}$,
Yilin Wang$^{1}$,
Qiang Xu$^{3}$,
Luyao Dong$^{2}$,
Xing He$^{2,*}$,
Wen Wu$^{1,*}$
\\[0.5em]
\small
$^{1}$Shanghai Artificial Intelligence Laboratory
\quad
$^{2}$Shanghai Jiao Tong University
\\
\small
$^{3}$The Chinese University of Hong Kong
\\[0.5em]
\small
\texttt{zby112200@sjtu.edu.cn, xldai26@cse.cuhk.edu.hk, wangyilin@pjlab.org.cn}
\\
\small
\texttt{qxu@cse.cuhk.edu.hk, sjtu4911226@sjtu.edu.cn, xhe@sjtu.edu.cn}
\\
\small
\texttt{wuwen@pjlab.org.cn}
\\[0.3em]
\small
$^{*}$Corresponding authors
}

\date{}

\etocdepthtag.toc{main}

\iclrfinalcopy

\begin{document}

\maketitle

\begin{abstract}
Time Series Foundation Models (TSFMs) provide general-purpose
forecasting capabilities, yet their forecasts can remain misaligned
with the temporal characteristics of a target domain.
This motivates lightweight forecast correction even when only
black-box access to the TSFM is available.
Our key observation is that two forms of historical information
provide complementary correction signals: historical analogues
reveal how similar contexts evolved, while past forecast errors
reveal a TSFM's characteristic deviations in the target domain.
Based on this observation, we propose ECHO, a black-box
forecast correction framework that learns from both sources.
ECHO uses AnalogE to assess retrieved historical analogues
together with the current context and base forecast,
learning transferable corrections through cross-domain pretraining.
In parallel, RecentE learns target-specific residual corrections
from past forecast errors and refines them as causally available
feedback accumulates.
ECHO dynamically combines the two corrections according to
their observed forecasting performance.
Across 5 representative TSFMs and 9 real-world datasets,
ECHO reduces MSE in all 45 backbone--dataset settings,
yielding an average relative reduction of 9.26\% in
macro-average MSE.
Its gains also extend to long-horizon forecasting.
\end{abstract}

\section{Introduction}

Time series forecasting is widely used in real-world applications,
including energy management \citep{hong2016probabilistic},
traffic management \citep{vlahogianni2014short},
financial analysis \citep{fischer2018deep},
and industrial monitoring \citep{malhotra2015long}.
With advances in deep learning, time series forecasting has evolved
from classical statistical approaches \citep{winters1960forecasting,box2015time} to deep learning methods \citep{zhou2021informer,zeng2023transformers}.
Conventional deep learning models are typically trained on specific
datasets to capture domain-specific temporal patterns \citep{wu2021autoformer,zhang2026freqcycle}.
Inspired by large-scale pretraining in natural language processing \citep{devlin2019bert,brown2020language},
Time Series Foundation Models (TSFMs), such as TimesFM \citep{das2023decoder},
Chronos \citep{ansari2024chronos},
and Moirai \citep{woo2024unified}, are pretrained on multi-domain time series to support
general-purpose forecasting.
However, time series from different domains exhibit substantial
differences in temporal patterns and distributional properties
\citep{godahewa2021monash}.
Pretrained TSFMs may not fully accommodate this heterogeneity,
limiting their forecasting performance in specific domains
\citep{aksu2024gift}.
These limitations highlight the need for efficient target-domain
adaptation.

Leveraging information available in the target domain is an important
avenue for adapting TSFMs \citep{das2024context}.
Existing approaches leverage cross-variate dependencies
\citep{ekambaram2024tiny},
exogenous covariates \citep{arango2025chronosx},
or textual context \citep{chen2026trace}
to enrich the information available for forecasting.
Among the available information sources, historical observations
and past model forecasts can be accumulated during routine
forecasting, providing a direct basis for target-domain adaptation.
Prior work exploits this history by retrieving relevant historical
cases to provide references for how the current series may evolve
\citep{tire2024retrieval,zhang2025timeraf,ning2026ts}, or by using past forecasts and their
observed outcomes to improve subsequent predictions through
forecast combination or learned corrections \citep{lee2025lightweight,liang2026forecast}.
However, these history-based methods typically focus on one
form of historical information.
Historical analogues provide references for how similar contexts
evolved, while past forecast errors reveal how a particular
backbone deviates from observed outcomes in the target domain.
Their complementary roles motivate us to explore how to learn
effective corrections from these two distinct forms of historical
evidence and coordinate their contributions to the current forecast.

Building on this perspective, we propose ECHO, a framework
for correcting TSFM forecasts using historical analogues
and past forecast errors.
To account for their distinct properties, ECHO employs two
correction modules.
Historical Analog Evidence (AnalogE) uses historical analogues, comprising past contexts similar to the current context and their observed futures. By assessing these analogues together with the current context and base forecast, it constructs a historical reference and determines how strongly to adjust the base forecast toward it. These correction strategies are learned through cross-domain pretraining.
Adaptive Recent Evidence (RecentE) learns residual patterns
from past forecast errors, conditioned on the current context
and base forecast, to provide corrections tailored to the
target domain.
It is initialized using target-domain training data and
periodically refined as newly matured forecasting feedback
becomes available.
To coordinate the two corrections, Temporal Reliability
Aggregation (TempRA) estimates their reliability from historical
forecasting performance and adjusts their contributions
to the final forecast.
ECHO keeps the backbone frozen and accesses only its inputs
and forecast outputs, enabling black-box target-domain adaptation.

Our main contributions are summarized as follows:
\begin{itemize}[leftmargin=1.2em]

\item Complementary historical evidence for forecast correction.
We jointly exploit historical analogues and past forecast errors
for TSFM adaptation, bringing together references for temporal
evolution and evidence of model-specific errors in the target
domain to guide forecast correction.

\item An evidence-driven correction framework.
We propose ECHO, which learns transferable corrections from
historical analogues through cross-domain pretraining (AnalogE)
and target-specific residual corrections from past forecast
errors (RecentE).
It coordinates the two corrections using reliability estimates
derived from historical forecasting performance (TempRA),
enabling target-domain adaptation with a black-box TSFM.

\item Extensive empirical evaluation.
Across 5 representative TSFMs and 9 real-world datasets,
ECHO reduces MSE in all 45 backbone--dataset settings,
with an average relative reduction of 9.26\% in macro-average MSE.
These improvements also extend to long-horizon forecasting.

\end{itemize}

\section{Related Work}

\subsection{Time Series Foundation Models}

Time Series Foundation Models (TSFMs) aim to learn transferable
temporal knowledge through large-scale pretraining \citep{liu2024timer,goswami2024moment}.
Existing research explores both adapting pretrained language
models for time series forecasting and developing native
time series foundation models.
Language-model adaptation approaches include partial finetuning
\citep{zhou2023one}, numerical encoding \citep{gruver2023large},
domain instructions and cross-modal alignment
\citep{liu2024unitime}, and input reprogramming
\citep{jin2024time}.
Native TSFMs adopt diverse architectures and representations:
Chronos \citep{ansari2024chronos} uses discretized autoregressive
modeling, TimesFM \citep{das2023decoder} and
Timer \citep{liu2024timer} adopt patch-based representations,
and Moirai \citep{woo2024unified} supports unified modeling
across frequencies and variables.
Subsequent developments include multivariate and covariate-informed
forecasting in Chronos-2 \citep{ansari2025chronos},
flow-matching-based probabilistic forecasting in Sundial
\citep{liu2025sundial},
lightweight architectures in TTM \citep{ekambaram2024tiny},
sparse mixture-of-experts modeling in Moirai-MoE
\citep{liu2024moirai},
and long-context modeling in Timer-S1 \citep{liu2026timer}.
However, differences in temporal patterns and distributions
across domains can limit the performance of frozen TSFMs
in specific target environments, motivating target-domain
adaptation \citep{aksu2024gift}.

\subsection{Retrieval-Augmented Time Series Forecasting}

Retrieval-augmented time series forecasting retrieves relevant
historical sequences to provide additional context or reference
information for the current forecast \citep{ning2026ts,tire2024retrieval}.
Existing studies explore different ways to incorporate retrieved
information: ReTime~\citep{jing2022retrieval} investigates retrieval-based
forecasting,
RATD~\citep{liu2024retrieval} integrates retrieved information
into diffusion-based forecasting,
TimeRAG~\citep{yang2025timerag} augments language-model forecasting
contexts with retrieved sequences,
and RAFT~\citep{han2025retrieval} aggregates retrieved historical
futures across multiple temporal scales and combines them
with the current input through a linear predictor.
Recent studies further integrate retrieval mechanisms with TSFMs.
TimeRAF introduces learnable retrieval and channel prompting~\citep{zhang2025timeraf},
while TS-RAG combines retrieved representations with internal
TSFM representations~\citep{ning2026ts}.
These integration mechanisms require interaction with internal
backbone representations,
limiting their direct applicability to black-box TSFMs.
Naive RAF~\citep{tire2024retrieval} supports black-box forecasting
by augmenting the input context with retrieved examples, but its gains depend
on the backbone's ability to exploit this information.
Learning to use historical analogues to correct forecasts
without accessing backbone internals therefore merits
further study.

\subsection{Forecast Adaptation and Correction}

Beyond historical sequences, past forecasts and their observed
outcomes provide another source of information for adaptation.
Existing adaptation methods differ in their access requirements
to the backbone.
SOLID~\citep{chen2024calibration} finetunes the prediction layer
using contextually similar samples, while
DSOF~\citep{lau2025fast} updates forecasting models through
fast and slow online learning streams.
TAFAS~\citep{kim2025battling} and Ada-X~\citep{liang2026forecast}
keep the backbone frozen, but optimizing their input-side modules
still requires backpropagation through the backbone.
Black-box approaches instead improve forecasts through external
combination or correction without accessing backbone internals.
ELF~\citep{lee2025lightweight} combines a lightweight online forecaster with foundation-model
predictions; Ada-Y~\citep{liang2026forecast}, the output-side variant of $\delta$-Adapter,
learns corrections from base forecasts; and ORCA~\citep{dai2026learning} learns residual
corrections conditioned on both the input context and the base
forecast.

Taken together, these studies demonstrate the value of historical
analogues and forecasting feedback for adaptation, but generally
study them separately.
How to learn corrections tailored to the distinct properties
of these two forms of evidence and coordinate their contributions
in black-box TSFMs warrants further investigation.
\section{Methodology}
\begin{figure*}[t]
    \centering
    \includegraphics[width=\textwidth]{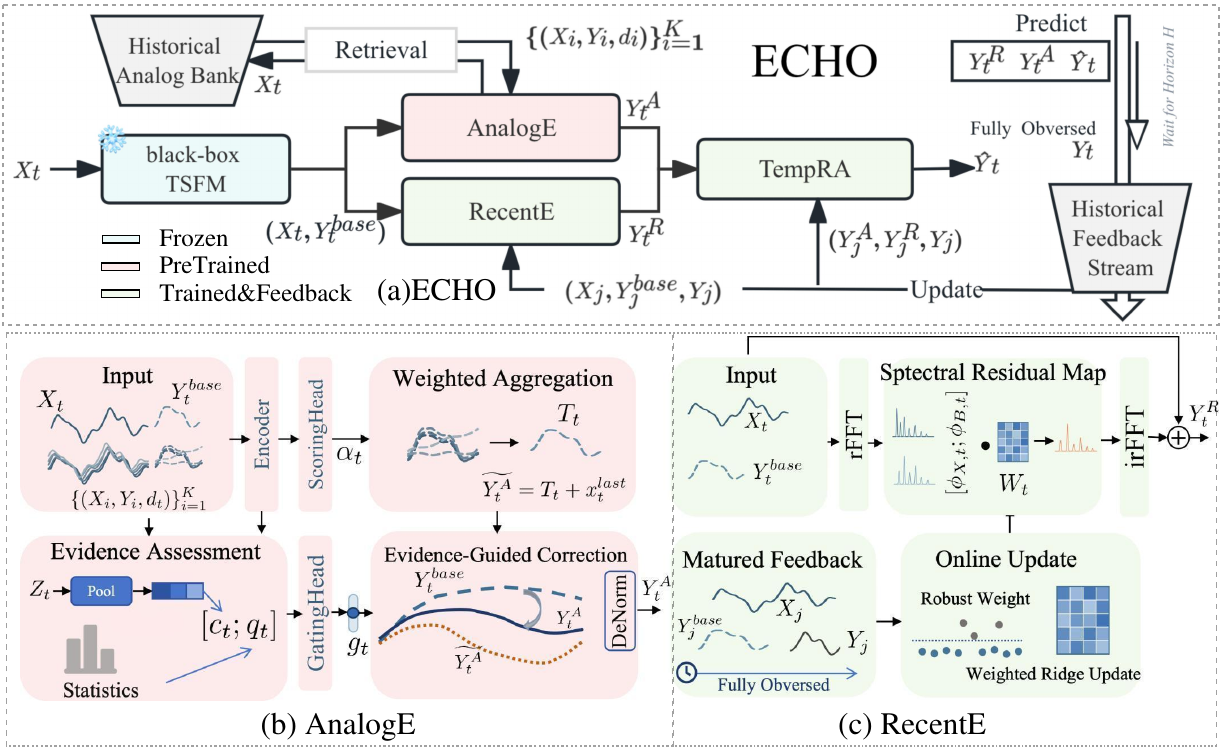}
    \caption{
    Overview of ECHO.
    (a) Overall framework.
    (b) AnalogE constructs a weighted historical template and estimates
    a gate to control the correction of the base forecast.
    (c) RecentE predicts spectral residuals and updates its residual
    map using fully observed feedback.
    }
    \label{fig:echo_framework}
\end{figure*}

Figure~\ref{fig:echo_framework} presents an overview of ECHO.
AnalogE and RecentE learn corrections from historical analogues
and past forecast errors, respectively, while TempRA coordinates
the two corrections based on their historical forecasting performance.

\subsection{Problem Formulation}

We consider target-domain forecasting with a black-box TSFM.
For time series with $D$ variates, forecast correction is performed
independently for each variate.
For clarity, we present the formulation for a single variate
and omit the variate index.
At each forecast origin $t$, observations up to and including
time $t$ are available.
The input $X_t\in\mathbb{R}^{L}$ contains the latest $L$
observations, and the target $Y_t\in\mathbb{R}^{H}$ covers
the next $H$ time steps, $t+1,\ldots,t+H$.
Let $Y_t^{\mathrm{base}}\in\mathbb{R}^{H}$ denote the base
forecast produced by the TSFM for this variate.

Let $\mathcal{H}_t$ denote the historical evidence available
for this variate at forecast origin $t$.
An external correction framework $G$ uses this evidence
to refine the base forecast:
\begin{equation}
\hat{Y}_t
=
G\!\left(
X_t,
Y_t^{\mathrm{base}};
\mathcal{H}_t
\right).
\end{equation}

\subsection{Historical Analog Evidence}

Historical Analog Evidence (AnalogE) corrects base forecasts
using retrieved historical analogues.
It constructs a historical reference and estimates a gate
that controls how strongly the base forecast is adjusted
toward this reference.

Given $X_t$, AnalogE retrieves $K$ context--future pairs
$(X_i,Y_i)$ with fully observed futures from a fixed training-set
memory, along with their retrieval distances $d_i$.
Retrieval details are provided in Appendix~\ref{app:Retrieval}.
AnalogE operates on window-normalized sequences, and corrected
forecasts are mapped back to the dataset-standardized scale.
We subtract the last observations
$x_i^{\mathrm{last}}$ and $x_t^{\mathrm{last}}$ of the corresponding
contexts from each historical future $Y_i$ and the base forecast
$Y_t^{\mathrm{base}}$, respectively, obtaining the historical
transitions $\Delta_i\in\mathbb{R}^{H}$ and the base forecast
transition $p_t^{\mathrm{base}}\in\mathbb{R}^{H}$.

\paragraph{Historical Template Construction.}
Under non-stationarity, similar contexts may lead to different
futures, making retrieval distances alone insufficient to assess
the usefulness of historical candidates~\citep{zhou2026stationarity}.
AnalogE therefore reassesses retrieved context--future pairs
jointly with the current context and base forecast.

An encoder integrates this information and models relationships
among candidates:
\begin{equation}
Z_t
=
\operatorname{Encoder}
\left(
X_t,
p_t^{\mathrm{base}},
\{(X_i,\Delta_i)\}_{i=1}^{K},
\mathbf d_t
\right),
\end{equation}
where $\mathbf d_t=(d_1,\ldots,d_K)$ and
$Z_t\in\mathbb{R}^{K\times d_z}$ contains the candidate
representations, with $Z_{t,i}$ denoting the $d_z$-dimensional
representation of candidate $i$.

The resulting representations $Z_t$ and retrieval distances
$\mathbf d_t$ are passed to a trainable scoring head, which
refines distance-based scores through bounded adjustments and
internally applies softmax across the $K$ candidates.
Its output weights are used to construct the historical template:
\begin{equation}
\boldsymbol{\alpha}_t
=
\operatorname{ScoringHead}(Z_t,\mathbf d_t),
\qquad
T_t=\sum_{i=1}^{K}\alpha_{t,i}\Delta_i,
\end{equation}
where $T_t\in\mathbb{R}^{H}$ aggregates historical transitions
according to their estimated usefulness for the current forecast.

\paragraph{Evidence Gate Estimation.}
Disagreement between the historical reference and the base
forecast does not necessarily warrant a substantial correction.
AnalogE pools representations across candidates to obtain
a candidate summary representation $c_t\in\mathbb{R}^{d_z}$
and concatenates it with explicit statistics
$q_t\in\mathbb{R}^{d_q}$ to form the gating input $e_t$.
These statistics characterize historical support quality,
correction evidence ambiguity, and foundation--experience
disagreement.
Their definitions are provided in Appendix~\ref{app:evidence_statistics}.

The gating head estimates correction strength from $e_t$:
\begin{equation}
g_t
=
\operatorname{GatingHead}(e_t)
=
g_{\max}\sigma\left(
\operatorname{MLP}(e_t)
\right),
\end{equation}
where $\sigma$ is the sigmoid function and $g_{\max}$ bounds
the gate $g_t$.

\paragraph{Evidence-Guided Correction.}
AnalogE aligns the historical transition template $T_t$ with
the latest observation $x_t^{\mathrm{last}}$.
The gate $g_t$ controls how far the base forecast moves toward
this aligned reference, yielding the corrected forecast $Y_t^{A}$:
\begin{equation}
Y_t^{A}
=
Y_t^{\mathrm{base}}
+
g_t\left(x_t^{\mathrm{last}}+T_t-Y_t^{\mathrm{base}}\right).
\end{equation}
Through cross-domain pretraining, AnalogE learns to construct
historical templates and estimate correction gates that jointly
guide forecast correction.

\subsection{Adaptive Recent Evidence}

Adaptive Recent Evidence (RecentE) learns residual patterns
in the target environment to correct base forecasts and
refines its corrections as new forecasting feedback becomes available.
For a historical instance $j$ with a fully observed future,
the residual target is $r_j=Y_j-Y_j^{\mathrm{base}}$.

\paragraph{Spectral Residual Adaptation.}
RecentE estimates residuals conditioned on the current context
$X_t$ and base forecast $Y_t^{\mathrm{base}}$.
We center $X_t$ by its mean and $Y_t^{\mathrm{base}}$ by the
latest context observation, transform both into the frequency
domain, and concatenate their spectra into
$\phi_t\in\mathbb{C}^{n_L+n_H}$.
Here, $n_L=\lfloor L/2\rfloor+1$ and
$n_H=\lfloor H/2\rfloor+1$ are the corresponding numbers
of frequency coefficients.

A complex linear map
$W_t\in\mathbb{C}^{(n_L+n_H)\times n_H}$
predicts the residual spectrum:
\begin{equation}
\hat z_t=\phi_tW_t.
\end{equation}
The predicted spectrum is transformed back to a length-$H$
residual sequence and added to the base forecast:
\begin{equation}
Y_t^R
=
Y_t^{\mathrm{base}}
+
\operatorname{irFFT}(\hat z_t,n=H).
\end{equation}

\paragraph{Residual Map Fitting and Refinement.}
RecentE initializes the residual map on training instances
and periodically updates it using selected mature records
from the feedback stream $\mathcal M_t$ defined in
Section~\ref{sec:tempra}.
The map is fitted to residual target spectra through
weighted ridge regression, alternating between estimating
robust sample weights from fitting errors and refitting
the map with these weights fixed.
This procedure limits the influence of large fitting errors.
The fitting objective and online update details are provided
in Appendix~\ref{app:recent_updates}.

\subsection{Temporal Reliability Aggregation}
\label{sec:tempra}

Temporal Reliability Aggregation (TempRA) assesses the
reliability of AnalogE and RecentE from historical forecasting
performance and adjusts their contributions to the final forecast.

At forecast origin $t$, the available historical feedback forms the following record set:
\begin{equation}
\mathcal M_t
=
\left\{
\left(
X_j,
Y_j^{\mathrm{base}},
Y_j^A,
Y_j^R,
Y_j
\right)
\;\middle|\;
j+H\leq t
\right\}.
\end{equation}
The condition $j+H\leq t$ ensures that each record's target horizon is fully observed.
From this feedback stream, RecentE draws historical inputs, base forecasts,
and observed outcomes $(X_j,Y_j^{\mathrm{base}},Y_j)$ for residual learning,
while TempRA draws the two corrected forecasts and their observed outcomes
$(Y_j^A,Y_j^R,Y_j)$ for reliability estimation.

TempRA constructs three temporal views from this feedback stream.
The long view summarizes accumulated forecasting performance,
while the short view uses a recent subset of sparsely selected
feedback records.
The recent view uses more densely sampled feedback to capture
local performance differences.
Within each view, TempRA computes prediction errors on the selected records
and uses these errors to assign AnalogE a weight $w_t^v\in[0,1]$,
where $v\in\{L,S,R\}$.
The corresponding RecentE weight is $1-w_t^v$.
View-specific record selection and weight estimation are detailed
in Appendix~\ref{app:tempra_details}.

We combine the three weights and aggregate the two forecasts:
\begin{equation}
w_t=\lambda_Lw_t^L+\lambda_Sw_t^S+\lambda_Rw_t^R,
\qquad
\hat Y_t=w_tY_t^A+(1-w_t)Y_t^R ,
\end{equation}
where $\lambda_v\geq0$ and $\sum_v\lambda_v=1$.

\section{Experimental Setup}

\paragraph{Datasets and Data Usage.}
We evaluate ECHO on nine time series datasets:
ETTh1, ETTh2, ETTm1, and ETTm2~\citep{zhou2021informer};
Exchange Rate, Electricity, and Solar-Energy~\citep{lai2018modeling};
Weather~\citep{wu2021autoformer};
and PEMS08~\citep{song2020spatial}.
These datasets cover electricity, exchange rates, weather, solar power
generation, and traffic flow, with diverse sampling frequencies and temporal dynamics. Following TS-RAG~\citep{ning2026ts}, we pretrain AnalogE using queries
sampled from the Chronos pretraining corpus~\citep{ansari2024chronos}
and keep it frozen on all downstream datasets.

\paragraph{Forecasting Backbones.}
We evaluate ECHO with five TSFMs:
TimesFM 2.5~\citep{das2023decoder},
Chronos-2~\citep{ansari2025chronos},
Chronos-Bolt~\citep{ansari2024chronos},
Moirai 2.0~\citep{liu2025moirai} and Sundial~\citep{liu2025sundial}.
All backbones remain frozen throughout evaluation.

\paragraph{Evaluation Protocol and Implementation.}

Unless otherwise specified, experiments use TimesFM 2.5
as the frozen backbone, with a context length of $L=512$,
a forecast horizon of $H=64$, and $K=20$ retrieved cases.
Online updates use only feedback whose complete target horizon is observed
by the current forecast origin.
We use Mean Squared Error (MSE) as the primary evaluation metric.
Mean Absolute Error (MAE) results are reported in
Appendix~\ref{app:additional_results}.
Additional experimental details are provided in
Appendix~\ref{app:experimental_details}.

\section{Main Results and Model Analysis}
We evaluate ECHO's forecasting performance, examine the effectiveness
of its components and learning designs, and assess its computational
efficiency.

\subsection{Main Results}
\paragraph{Cross-Backbone Forecasting.}
To evaluate the generality of ECHO as a forecast correction framework,
we compare the original and ECHO-corrected forecasts of 5 frozen
TSFMs on 9 datasets.
As shown in Table~\ref{tab:cross_backbone}, ECHO achieves lower MSE in all 45 dataset--backbone
configurations, with relative changes in macro-average MSE
ranging from -14.43\% to -5.55\%.
These results support the applicability of ECHO across TSFMs
with different architectures and forecasting behaviors.

\begin{table*}[t]
\caption{
Cross-backbone forecasting MSE on 9 datasets with context length
$L=512$ and forecast horizon $H=64$.
Ori denotes the original forecasts of each frozen TSFM,
and ECHO denotes its corrected forecasts.
The better result within each pair is bolded.
Overall reports the relative change in macro-average MSE
over the 9 datasets.
}
\label{tab:cross_backbone}
\centering

\fontsize{8pt}{7.5pt}\selectfont
\setlength{\tabcolsep}{5.5pt}
\renewcommand{\arraystretch}{1.08}

\begin{tabular}{lcccccccccc}
\toprule

\textbf{Models}
& \multicolumn{2}{c}{\textbf{TimesFM 2.5}}
& \multicolumn{2}{c}{\textbf{Chronos-2}}
& \multicolumn{2}{c}{\textbf{Chronos-Bolt}}
& \multicolumn{2}{c}{\textbf{Moirai 2.0}}
& \multicolumn{2}{c}{\textbf{Sundial}}
\\

\cmidrule(lr){2-3}
\cmidrule(lr){4-5}
\cmidrule(lr){6-7}
\cmidrule(lr){8-9}
\cmidrule(lr){10-11}

\textbf{MSE}
& Ori & ECHO
& Ori & ECHO
& Ori & ECHO
& Ori & ECHO
& Ori & ECHO
\\

\midrule

ETTh1
& .355 & \textbf{.346}
& .377 & \textbf{.348}
& .362 & \textbf{.343}
& .367 & \textbf{.340}
& .337 & \textbf{.336}
\\

ETTh2
& .257 & \textbf{.242}
& .266 & \textbf{.246}
& .252 & \textbf{.240}
& .266 & \textbf{.241}
& .242 & \textbf{.233}
\\

ETTm1
& .280 & \textbf{.263}
& .309 & \textbf{.274}
& .311 & \textbf{.271}
& .322 & \textbf{.273}
& .268 & \textbf{.257}
\\

ETTm2
& .148 & \textbf{.136}
& .149 & \textbf{.138}
& .149 & \textbf{.139}
& .155 & \textbf{.138}
& .149 & \textbf{.135}
\\

Exchange
& .059 & \textbf{.056}
& .060 & \textbf{.058}
& .067 & \textbf{.064}
& .060 & \textbf{.058}
& .082 & \textbf{.067}
\\

Weather
& .135 & \textbf{.124}
& .143 & \textbf{.128}
& .152 & \textbf{.134}
& .148 & \textbf{.127}
& .136 & \textbf{.123}
\\

Electricity
& .120 & \textbf{.116}
& .116 & \textbf{.113}
& .113 & \textbf{.111}
& .117 & \textbf{.114}
& .107 & \textbf{.106}
\\

PEMS08
& .097 & \textbf{.089}
& .232 & \textbf{.179}
& .289 & \textbf{.190}
& .112 & \textbf{.098}
& .094 & \textbf{.088}
\\

Solar
& .208 & \textbf{.180}
& .147 & \textbf{.145}
& .469 & \textbf{.359}
& .181 & \textbf{.159}
& .210 & \textbf{.189}
\\

\midrule

\rowcolor{gray!12}
\textbf{Overall}
& -- & \imp{-6.60\%}
& -- & \imp{-9.36\%}
& -- & \imp{-14.43\%}
& -- & \imp{-10.38\%}
& -- & \imp{-5.55\%}
\\

\bottomrule
\end{tabular}
\end{table*}

\paragraph{Long-Horizon Forecasting.}
We next examine whether ECHO's improvements extend to longer forecast
horizons by comparing original and ECHO-corrected forecasts on
9 datasets with $H\in\{96,192,336,720\}$.
For horizons beyond AnalogE's fixed training length, we apply retrieval
and correction in a rolling manner without retraining AnalogE.
As shown in Table~\ref{tab:long_horizon}, ECHO achieves lower MSE
in most dataset--horizon configurations.
The relative change in macro-average MSE decreases from
-6.91\% at $H=96$ to -8.50\% at $H=720$,
supporting the effectiveness of ECHO over longer forecast horizons.

\begin{table*}[t]
\caption{
Long-horizon MSE of TimesFM 2.5 with $L=512$ and
$H\in\{96,192,336,720\}$.
ETTh Avg.\ and ETTm Avg.\ average the corresponding ETT datasets.
Lower is better; pairwise best results are bolded.
Overall reports the relative change in macro-average MSE
over the nine original datasets.
}
\label{tab:long_horizon}
\centering

\fontsize{8pt}{8.5pt}\selectfont
\setlength{\tabcolsep}{0.9pt}
\renewcommand{\arraystretch}{1.02}

\begin{tabular}{@{}c@{\hspace{0.35em}}c@{}}

\begin{tabular}[t]{lcccccccc}
\toprule
\textbf{Dataset}
& \multicolumn{2}{c}{\textbf{96}}
& \multicolumn{2}{c}{\textbf{192}}
& \multicolumn{2}{c}{\textbf{336}}
& \multicolumn{2}{c}{\textbf{720}}
\\
\cmidrule(lr){2-3}
\cmidrule(lr){4-5}
\cmidrule(lr){6-7}
\cmidrule(lr){8-9}
& Ori & ECHO
& Ori & ECHO
& Ori & ECHO
& Ori & ECHO
\\
\midrule

ETTh Avg.
& .337 & \textbf{.320}
& .391 & \textbf{.367}
& .413 & \textbf{.390}
& .412 & \textbf{.399}
\\

ETTm Avg.
& .238 & \textbf{.222}
& .293 & \textbf{.273}
& .340 & \textbf{.322}
& .422 & \textbf{.400}
\\

Exchange
& .091 & \textbf{.085}
& .194 & \textbf{.181}
& .360 & \textbf{.326}
& 1.046 & \textbf{.944}
\\

Weather
& .159 & \textbf{.144}
& .209 & \textbf{.189}
& .275 & \textbf{.246}
& .362 & \textbf{.332}
\\

\bottomrule
\end{tabular}
&
\begin{tabular}[t]{lcccccccc}
\toprule
\textbf{Dataset}
& \multicolumn{2}{c}{\textbf{96}}
& \multicolumn{2}{c}{\textbf{192}}
& \multicolumn{2}{c}{\textbf{336}}
& \multicolumn{2}{c}{\textbf{720}}
\\
\cmidrule(lr){2-3}
\cmidrule(lr){4-5}
\cmidrule(lr){6-7}
\cmidrule(lr){8-9}
\textit{cont.}
& Ori & ECHO
& Ori & ECHO
& Ori & ECHO
& Ori & ECHO
\\
\midrule

PEMS08
& .109 & \textbf{.098}
& .143 & \textbf{.126}
& .177 & \textbf{.156}
& .249 & \textbf{.199}
\\

Electricity
& .131 & \textbf{.127}
& .148 & \textbf{.146}
& \textbf{.165} & .166
& \textbf{.205} & .209
\\

Solar
& .222 & \textbf{.193}
& .255 & \textbf{.231}
& .292 & \textbf{.230}
& .315 & \textbf{.235}
\\

\rowcolor{gray!12}
\textbf{Overall}
& \multicolumn{2}{c}{\imp{-6.91\%}}
& \multicolumn{2}{c}{\imp{-7.11\%}}
& \multicolumn{2}{c}{\imp{-8.10\%}}
& \multicolumn{2}{c}{\imp{-8.50\%}}
\\

\bottomrule
\end{tabular}

\end{tabular}
\end{table*}
\subsection{Complementarity of Evidence Sources}

Historical and recent evidence offer complementary strengths,
motivating their joint use for robust forecast correction.
To assess this, we compare three configurations:
AnalogE-only, which corrects the base forecast using only historical
evidence from AnalogE; RecentE-only, which corrects the base forecast
using only forecast-error evidence from RecentE; and full ECHO,
which combines both sources.

\begin{figure*}[t]
    \centering
    \begin{minipage}[t]{0.47\textwidth}
        \vspace{0pt}
        \centering
        \begin{minipage}[t][5.8cm][t]{\linewidth}
            \vspace{0pt}
            \centering
            \includegraphics[width=\linewidth]
            {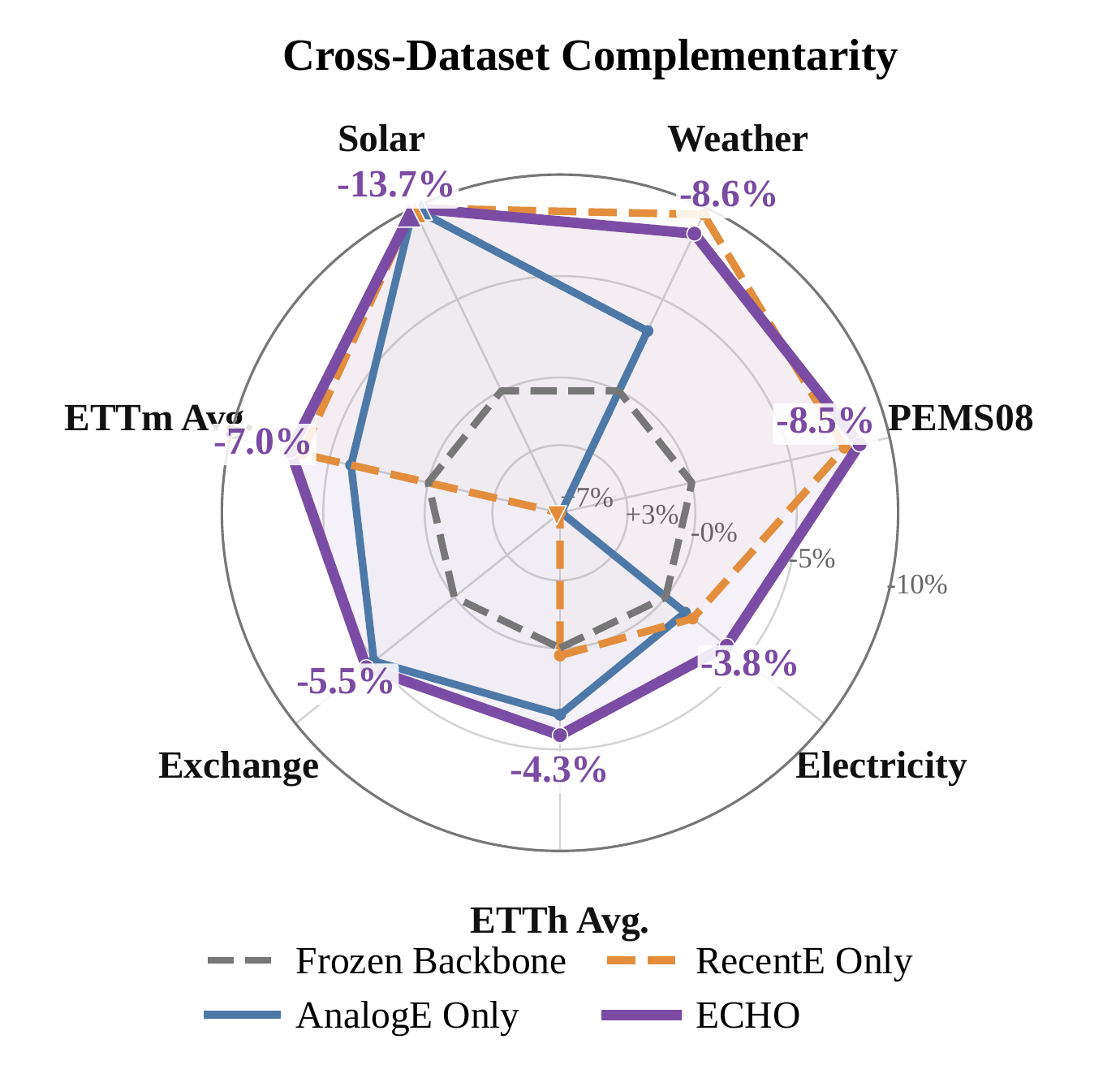}
        \end{minipage}
        \par
        \caption{
            Relative MSE changes (\%) of AnalogE-only, RecentE-only,
            and ECHO against TimesFM 2.5 across 9 datasets,
            averaging ETT pairs.
            Larger reductions appear farther outward. Extreme values are clipped for visualization and marked with triangles.
        }
        \label{fig:evidence_radar}
    \end{minipage}
    \hfill
    \begin{minipage}[t]{0.52\textwidth}
        \vspace{0pt}
        \centering
        \begin{minipage}[t][5.8cm][t]{\linewidth}
            \vspace{0pt}
            \centering
            \includegraphics[width=\linewidth]
            {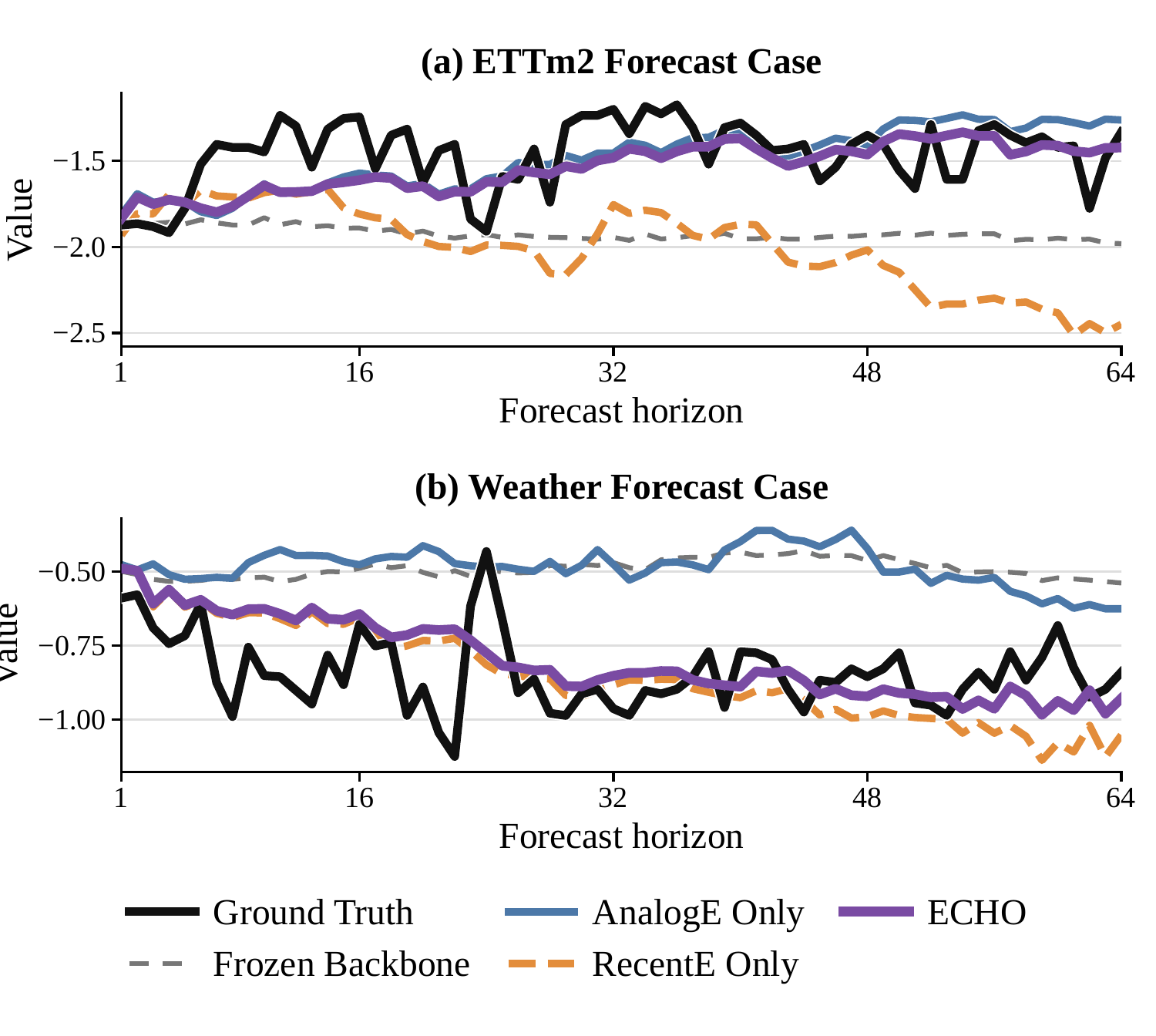}
        \end{minipage}
        \par
\caption{
Forecasting examples on (a) ETTm2 and (b) Weather.
AnalogE-only better captures the sequence level in (a),
while RecentE-only better captures the downward shift in (b).
ECHO combines the complementary corrections from both evidence sources.
}
        \label{fig:evidence_cases}
    \end{minipage}
\end{figure*}

Figure~\ref{fig:evidence_radar} summarizes relative MSE changes
against the frozen backbone across nine datasets.
Larger reductions appear farther outward.
AnalogE-only and RecentE-only exhibit different strengths across
datasets, while ECHO achieves more consistent improvements,
supporting the benefit of using both evidence sources jointly
(Appendix~\ref{app:complementarity}).

Figure~\ref{fig:evidence_cases} illustrates the complementary
strengths of the two evidence sources.
In Figure~\ref{fig:evidence_cases}(a), AnalogE-only better
captures the overall sequence level, whereas RecentE-only
increasingly underestimates the target values.
In Figure~\ref{fig:evidence_cases}(b), RecentE-only better
captures the downward shift, whereas AnalogE-only
overestimates the sequence level.

Using identical branch forecasts, TempRA also achieves lower
macro-average MSE than validation-selected fixed fusion (Appendix~\ref{app:fixed_fusion}).

\subsection{Role of the Base Forecast}
\begin{figure*}[t]
    \centering
    \includegraphics[width=\textwidth]
    {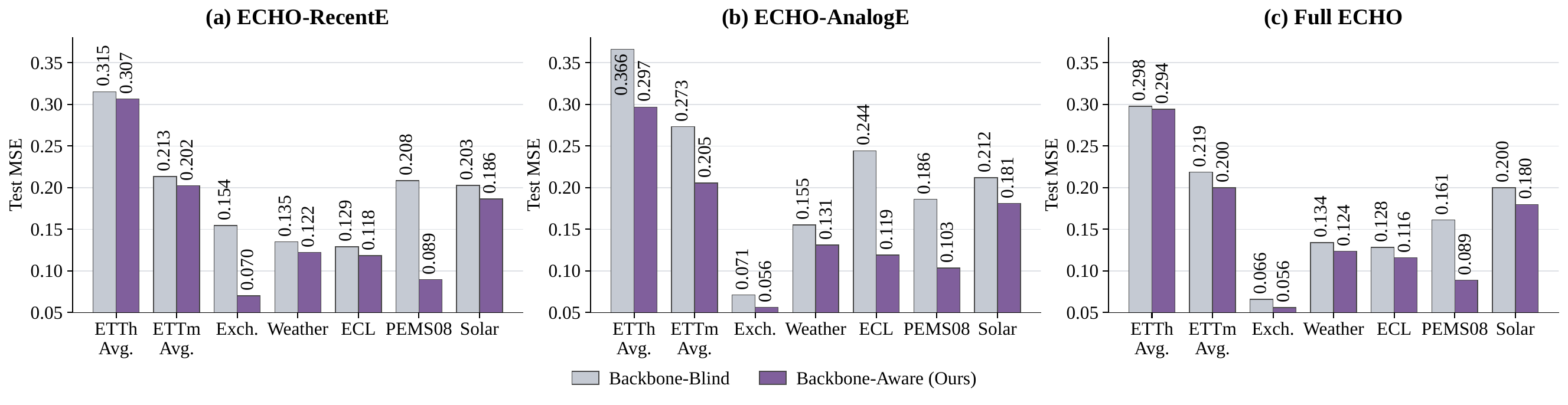}
\caption{
Role of the base forecast.
Backbone-Blind predicts directly without the base forecast.
Panels (a) and (b) compare the recent and historical analogue
branches.
Panel (c) compares ECHO.
ECL denotes Electricity.
Lower MSE indicates better forecasting performance.
}
    \label{fig:evidence_information_access}
\end{figure*}

We examine whether evidence is more effective for correcting
the backbone forecast than for independent prediction.
We construct Backbone-Aware and Backbone-Blind variants of
AnalogE, RecentE, and full ECHO.
For AnalogE, the aware variant corrects the backbone forecast
using retrieved cases, while the blind variant predicts directly
from them.
For RecentE, the aware variant predicts residual corrections,
while the blind variant predicts target values from the context.
For full ECHO, both variants use TempRA to aggregate the outputs
of AnalogE and RecentE: corrected forecasts in the aware variant
and independent predictions in the blind variant.
As shown in Figure~\ref{fig:evidence_information_access},
the Backbone-Aware variants outperform their Backbone-Blind
counterparts for both individual branches and full ECHO.

\subsection{Evidence Construction}

We evaluate AnalogE and RecentE against alternative methods
for using the corresponding historical evidence.
Full ECHO is also included as a reference.
For historical analog evidence, we compare AnalogE with
a black-box variant of TS-RAG~\citep{ning2026ts} and
Naive RAF~\citep{tire2024retrieval}.
Since the original TS-RAG fuses retrieved representations
with internal TSFM representations before the output projection,
we implement a variant compatible with black-box access (BB-TS-RAG).
For RAF, we use Naive RAF with the backbone kept frozen.
Table~\ref{tab:evidence_construction}(a) shows that AnalogE
outperforms both retrieval baselines in 3 of the 4 dataset groups,
supporting its effectiveness in constructing corrections
from historical analogues.

For adaptive recent evidence, we compare RecentE with
ELF~\citep{lee2025lightweight} and Ada-Y~\citep{liang2026forecast}.
ELF retains its original forecaster and dynamic ensemble weighting.
We additionally evaluate ELF+$Y^{\mathrm{base}}$ and Ada-Y+$X$,
which only add the base forecast and context to their respective
predictor inputs, leaving the remaining components unchanged.
These comparisons examine correction designs with matched
predictor inputs.
Table~\ref{tab:evidence_construction}(b) shows that RecentE
achieves lower MSE than all four alternatives on ETTm and Weather,
including those with the same predictor inputs.
These results support the effectiveness of RecentE's design beyond input access alone.
Detailed baseline configurations and additional results
are provided in Appendices~\ref{app:baseline_configs},
\ref{app:analog_construction}, and~\ref{app:recent_construction}.
\begin{table*}[t]
\caption{
MSE comparison of (a) historical and (b) recent evidence methods
with TimesFM 2.5.
$X$ and $Y^{\mathrm{base}}$ indicate predictor inputs.
ECHO is included as a full-framework reference.
ETTh and ETTm average their respective dataset pairs;
Exch.\ denotes Exchange Rate. Lower is better.
}
\label{tab:evidence_construction}
\centering
\footnotesize

\noindent\makebox[\textwidth][c]{%
\begin{minipage}[t]{0.42\textwidth}
\vspace{0pt}
\centering
\textbf{(a) Historical Analog Evidence}\par
\vspace{0.35em}

\setlength{\tabcolsep}{2.0pt}
\renewcommand{\arraystretch}{1.35}

\begin{tabular}{
@{}
>{\raggedright\arraybackslash}p{0.29\linewidth}
*{4}{>{\centering\arraybackslash}p{
\dimexpr(0.71\linewidth-8\tabcolsep)/4\relax
}}
@{}
}
\toprule
\textbf{Method}
& \textbf{ETTh}
& \textbf{ETTm}
& \textbf{Exch.}
& \textbf{Weather}
\\
\midrule

Backbone
& .306 & .214 & .059 & .135
\\
\midrule

BB-TS-RAG
& .299 & .212 & .074 & .139
\\

Naive RAF
& .299 & .213 & .058 & .130
\\
\midrule

AnalogE
& .297 & .205 & .056 & .131
\\
\midrule

\rowcolor{gray!10}
ECHO
& .294 & .200 & .056 & .124
\\
\bottomrule
\end{tabular}
\end{minipage}%
\hspace{0.02\textwidth}%
\begin{minipage}[t]{0.54\textwidth}
\vspace{0pt}
\centering
\textbf{(b) Adaptive Recent Evidence}\par
\vspace{0.35em}

\setlength{\tabcolsep}{1.5pt}
\renewcommand{\arraystretch}{1.15}

\begin{tabular}{
@{\hspace{0.04\linewidth}}
>{\raggedright\arraybackslash}p{0.22\linewidth}
>{\centering\arraybackslash}p{0.04\linewidth}
>{\centering\arraybackslash}p{0.09\linewidth}
*{4}{>{\centering\arraybackslash}p{
\dimexpr(0.59\linewidth-12\tabcolsep)/4\relax
}}
@{}
}
\toprule
\textbf{Method}
& $X$
& $Y^{\mathrm{base}}$
& \textbf{ETTh}
& \textbf{ETTm}
& \textbf{Exch.}
& \textbf{Weather}
\\
\midrule

ELF
& \checkmark & --
& .295 & .211 & .067 & .133
\\

ELF + $Y^{\mathrm{base}}$
& \checkmark & \checkmark
& .296 & .212 & .067 & .133
\\
\midrule

Ada-Y
& -- & \checkmark
& .307 & .214 & .059 & .134
\\

Ada-Y+$X$
& \checkmark & \checkmark
& .308 & .214 & .057 & .133
\\
\midrule

RecentE
& \checkmark & \checkmark
& .307 & .202 & .070 & .122
\\
\midrule

\rowcolor{gray!10}
ECHO
& \checkmark & \checkmark
& .294 & .200 & .056 & .124
\\
\bottomrule
\end{tabular}
\end{minipage}%
}
\end{table*}

\subsection{AnalogE Pretraining and RecentE Online Updates}
\begin{figure*}[t]
    \centering
    \includegraphics[width=\textwidth]{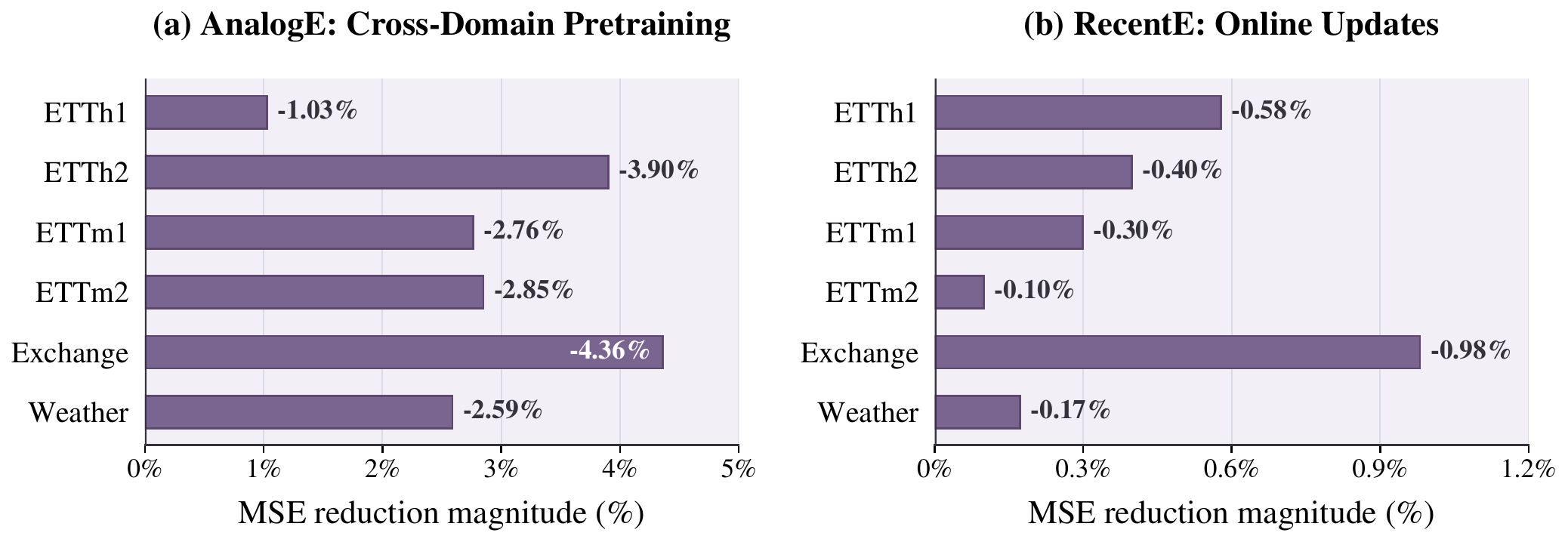}
    \caption{
    Effects of AnalogE Pretraining and RecentE Online Updates.
    (a) Relative MSE changes from cross-domain pretraining of AnalogE
    compared with training only on the target domain.
    (b) Relative MSE changes from online updates of RecentE
    compared with keeping the residual map frozen.
    Bar lengths represent relative reduction magnitudes.
    }
    \label{fig:learning_strategies}
\end{figure*}

We find that pretraining AnalogE on a multi-domain
dataset enables it to better use retrieved historical cases for
forecast correction than training only on target-domain data.
To assess this benefit, we compare the same AnalogE architecture
under these two training settings.
As shown in Figure~\ref{fig:learning_strategies}(a),
cross-domain pretraining improves AnalogE on all evaluated
datasets.

For RecentE, we examine whether online updates with matured
forecasting feedback help track error patterns specific to
the foundation model and target environment.
We compare continued online updates with a frozen residual map,
both starting from the same target-domain initialization.
Figure~\ref{fig:learning_strategies}(b) shows that online updates
yield small but consistent gains, supporting the benefit of
refining residual corrections as new feedback becomes available.

\subsection{Efficiency Analysis}  
We evaluate ECHO's parameter requirements and computational cost. AnalogE uses approximately 42K parameters shared across series, while RecentE maintains a linear residual map for each series. On ETTh1, ECHO takes 31.66 ms per inference query,
compared with 26.01 ms for the frozen backbone,
corresponding to an additional 5.65 ms. Each online update takes 486.69 ms, with this cost amortized over the predictions between successive updates. These results quantify the additional computational cost
of forecast correction.

\section{Conclusion}

We presented ECHO, a lightweight framework for correcting
black-box TSFM forecasts using historical analogues and
past forecast errors.
These sources provide complementary signals: how similar
contexts evolved and how the backbone deviates from observed
outcomes in the target domain.
ECHO learns transferable analogue-based corrections through
cross-domain pretraining and target-specific residual corrections
through online updates, dynamically combining them according
to observed forecasting performance.
Across 5 TSFMs and 9 real-world datasets, ECHO reduces MSE
in all 45 backbone--dataset settings, with an average relative
reduction of 9.26\% in macro-average MSE.
Its gains also extend to long-horizon forecasting.

\subsection*{AI Use Statement}
In this work, we used generative AI tools to assist with literature
search, writing portions of the code, and language polishing of
parts of the manuscript.
All AI-assisted content was reviewed and approved by the authors.
This included verifying references and their support for the
associated claims, checking code correctness, and ensuring that
the polished text accurately conveyed the intended meaning.
The authors take full responsibility for the final content of
this work, including text and code produced with the assistance
of generative AI tools.

\bibliography{iclr2027_conference}
\bibliographystyle{iclr2027_conference}

\appendix
\clearpage
\section*{Appendix}

\etocsettagdepth{main}{none}
\etocsettagdepth{appendix}{subsection}
\etocsettocstyle{\subsection*{Contents}}{}
\tableofcontents

\clearpage
\etocdepthtag.toc{appendix}

\section{Methodological Details}
\label{app:method_details}

This section provides additional methodological details of ECHO, covering historical case retrieval, explicit evidence statistics, causal residual updates, and temporal reliability aggregation.

\subsection{Historical Case Retrieval}
\label{app:Retrieval}

Given the current context $X_t$, we retrieve context--future
pairs $(X_i,Y_i)$ from the historical memory of the same variate.

We use a frozen encoder $f_{\mathrm{ret}}$ to map the query
and historical contexts into a shared embedding space,
and measure their similarity using squared Euclidean distance:
\begin{equation}
d_i
=
\left\|
f_{\mathrm{ret}}(X_t)-f_{\mathrm{ret}}(X_i)
\right\|_2^2.
\end{equation}
We select the $K$ nearest valid candidates as historical
evidence for AnalogE.
Historical futures $Y_i$ are excluded from retrieval keys
and are used only after retrieval to construct the correction
reference.
Invalid candidates are masked out; if no valid candidates
are available, AnalogE returns the base forecast.

\subsection{Explicit Evidence Statistics}
\label{app:evidence_statistics}

The evidence gate uses a 19-dimensional statistics vector $q_t$
to characterize historical support quality, correction evidence
ambiguity, and foundation--experience disagreement.
Table~\ref{tab:evidence_statistics} summarizes these groups.
All statistics use only the current context, retrieved historical
cases, and base forecast, without access to current target values.

\begin{table}[H]
\centering
\caption{Explicit evidence statistics used by AnalogE.
The three groups form the 19-dimensional vector $q_t$.}
\label{tab:evidence_statistics}
\small
\renewcommand{\arraystretch}{1.15}
\setlength{\tabcolsep}{4pt}
\begin{tabular}{@{}p{0.25\linewidth}cp{0.62\linewidth}@{}}
\toprule
\textbf{Category} & \textbf{Dim.} & \textbf{Statistics} \\
\midrule
Historical support quality
& 9
& Log-transformed minimum, weighted mean, weighted standard
deviation, and nearest-neighbor margin of retrieval distances;
weighted context RMSE, correlation, and last-value mismatch;
bounded context-volatility mismatch and signed log
context-volatility ratio.
\\
\midrule
Correction evidence ambiguity
& 6
& Normalized effective support; normalized top-4 template
discrepancy; absolute and relative transition dispersion;
directional agreement; and mean pairwise transition cosine.
\\
\midrule
Foundation--experience disagreement
& 4
& Mean absolute difference, symmetric relative RMS difference,
correlation, and maximum absolute difference between the
historical transition template and the base forecast transition.
\\
\bottomrule
\end{tabular}
\end{table}

Below, we clarify statistics whose computation is not fully
specified by their names.
We omit the forecast-origin index $t$ for readability.
The index $h$ denotes a forecast step, and subscripts $h$
indicate the corresponding vector components.
We use $\epsilon=10^{-6}$ for numerical stability,
$\operatorname{clip}_{[a,b]}$ to denote clipping to $[a,b]$,
and $\mathbf{1}[\cdot]$ to denote the indicator function.

\paragraph{Distance Weights and Reference Template.}
Let $\mathcal C$ denote the valid candidate set,
$d_i$ the retrieval distance of candidate $i$, and
$\Delta_i$ its historical transition vector.
The fixed distance weights and diagnostic template are
\begin{equation}
\beta_i
=
\frac{\exp(-d_i)}
{\sum_{j\in\mathcal C}\exp(-d_j)},
\qquad
T^{\mathrm{dist}}
=
\sum_{i\in\mathcal C}\beta_i\Delta_i.
\end{equation}
These weights use temperature 1 and are distinct from the
learned scoring-head weights $\alpha_{t,i}$.
The explicit statistics are computed using the diagnostic
template $T^{\mathrm{dist}}$, rather than the learned
template $T_t$.

The four distance statistics are the minimum distance,
the $\beta_i$-weighted mean and standard deviation,
and the nearest-neighbor margin.
The margin is the difference between the second-smallest
and smallest valid distances.
Each of these statistics is transformed by $\log(1+\cdot)$.
Context RMSE, correlation, and absolute last-value mismatch
are computed for each query--candidate pair and then averaged
using $\beta_i$.

\paragraph{Effective Support and Template Discrepancy.}
The normalized effective support is
\begin{equation}
s_{\mathrm{eff}}
=
\operatorname{clip}_{[0,1]}
\left(
\frac{(\sum_{i\in\mathcal C}\beta_i^2)^{-1}-1}{K-1}
\right).
\end{equation}
The normalization uses the configured number of candidate
slots $K$, even when fewer candidates are valid.

For a transition vector $u\in\mathbb R^H$, define its
stabilized root mean square (RMS) as
\begin{equation}
R(u)
=
\sqrt{\frac{1}{H}\sum_{h=1}^{H}u_h^2+\epsilon},
\end{equation}
where $H$ is the forecast horizon.
Let $T^{(4)}$ be the template formed from the four nearest
candidate slots, with distance weights renormalized over
their valid candidates.
The template discrepancy is
\begin{equation}
s_{\mathrm{top4}}
=
\frac{R(T^{(4)}-T^{\mathrm{dist}})}
{\max\!\left(R(T^{(4)})+R(T^{\mathrm{dist}}),\epsilon\right)}.
\end{equation}
Smaller values indicate greater consistency between the
top-4 template and the full retrieved template.

\paragraph{Transition Dispersion and Directional Agreement.}
Absolute and relative transition dispersion are
\begin{equation}
s_{\mathrm{abs}}
=
\sqrt{
\frac{1}{H}\sum_{h=1}^{H}
\sum_{i\in\mathcal C}
\beta_i(\Delta_{i,h}-T_h^{\mathrm{dist}})^2+\epsilon
},
\qquad
s_{\mathrm{rel}}
=
\frac{s_{\mathrm{abs}}}
{\max\!\left(\sum_{i\in\mathcal C}\beta_iR(\Delta_i),
\epsilon\right)}.
\end{equation}
Directional agreement measures the majority-sign fraction
at each forecast step, averaged over the horizon:
\begin{equation}
s_{\mathrm{dir}}
=
\frac{1}{H}\sum_{h=1}^{H}
\max\left(
\frac{1}{|\mathcal C|}
\sum_{i\in\mathcal C}\mathbf 1[\Delta_{i,h}>0],
\frac{1}{|\mathcal C|}
\sum_{i\in\mathcal C}\mathbf 1[\Delta_{i,h}<0]
\right).
\end{equation}
Here, $|\mathcal C|$ denotes the number of valid candidates.
Zero transitions contribute to neither sign.
Directional agreement and mean pairwise transition cosine
are unweighted over valid candidates.
The latter averages cosine similarity over distinct valid
candidate pairs and is set to zero when no such pair exists.

\paragraph{Context-Volatility Statistics.}
Volatility is the population standard deviation of
first differences.
Let $\sigma_X$ denote the query-context volatility,
$\sigma_i$ the volatility of historical context $X_i$,
and $\bar\sigma=\sum_{i\in\mathcal C}\beta_i\sigma_i$
the weighted historical-context volatility.
The bounded mismatch and signed log ratio are
\begin{equation}
s_{\mathrm{vol}}
=
\frac{|\bar\sigma-\sigma_X|}
{\max(\bar\sigma+\sigma_X,\epsilon)},
\qquad
s_{\mathrm{logvol}}
=
\operatorname{clip}_{[-8,8]}
\left(
\log\frac{\bar\sigma+\epsilon}{\sigma_X+\epsilon}
\right).
\end{equation}
The first statistic measures unsigned mismatch,
while the second preserves its direction.
Both compare historical and query contexts,
rather than future forecasts.

\paragraph{Relative Template--Backbone Disagreement.}
Let $p^{\mathrm{base}}$ denote the base forecast transition
defined in the main text, with the forecast-origin index omitted.
The symmetric relative RMS difference is
\begin{equation}
s_{\mathrm{dis}}
=
\frac{R(T^{\mathrm{dist}}-p^{\mathrm{base}})}
{\max\!\left(
R(T^{\mathrm{dist}})+R(p^{\mathrm{base}}),\epsilon
\right)}.
\end{equation}
The remaining disagreement statistics are the mean absolute
difference, centered correlation, and maximum absolute
difference between these two transition vectors.
\subsection{Causal Online Residual Updates}
\label{app:recent_updates}

RecentE and TempRA share the causal stream of matured feedback
$\mathcal M_t$, but apply different record selection rules.
For RecentE, feedback records are selected at intervals of $s$
forecast origins.
A selected record from origin $j$ becomes eligible for fitting
only when its complete target horizon has been observed,
i.e., $j+H\leq t$.
The residual map is refitted after every $u$ newly selected,
fully matured feedback records.
Fitting uses the initial training records and all selected
mature feedback available by time $t$, whose union is denoted
by $\mathcal D_t$.
Selected records are retained for subsequent updates.

Robust sample weights are computed from the fitting errors
of the current residual map.
For each record $i\in\mathcal D_t$, let $Y_i$ denote the
observed future, $Y_i^{\mathrm{base}}$ the frozen backbone's
forecast, and $\hat r_i$ the residual estimate produced
by the current map.
The fitting error is
\begin{equation}
e_i=(Y_i-Y_i^{\mathrm{base}})-\hat r_i.
\end{equation}

We partition the forecast horizon into $G=\lceil H/b\rceil$
contiguous groups, each containing at most $b$ time steps.
Let $\mathcal I_g$ denote the time-step indices in group $g$.
We compute the root mean squared error within each group
and average the resulting Huber weights to obtain
a sample weight:
\begin{equation}
q_{i,g}
=
\sqrt{
\frac{1}{|\mathcal I_g|}
\sum_{h\in\mathcal I_g}e_{i,h}^{2}
},
\qquad
a_i
=
\frac{1}{G}
\sum_{g=1}^{G}
\min\left(1,\frac{\delta}{q_{i,g}+\epsilon}\right).
\end{equation}
Here, $e_{i,h}$ is the fitting error at forecast step $h$,
$q_{i,g}$ is the root mean squared error in group $g$,
$\delta>0$ is the Huber threshold, and $\epsilon>0$
is a small constant for numerical stability.
Grouping is used only to compute the scalar sample weight
$a_i$; separate residual maps are not fitted for individual
groups.

With the sample weights fixed, the residual map is updated
through weighted ridge regression:
\begin{equation}
W_t
=
\arg\min_W
\sum_{i\in\mathcal D_t}
a_i\left\|\phi_iW-z_i\right\|_2^2
+
\lambda\|W\|_F^2.
\end{equation}
Here, $\phi_i$ contains the spectral input features derived
from the context and base forecast, and $z_i$ is the target
spectrum of the residual $Y_i-Y_i^{\mathrm{base}}$.
The complex linear map $W$ is optimized to obtain
the updated map $W_t$.
The norms $\|\cdot\|_2$ and $\|\cdot\|_F$ denote the
Euclidean and Frobenius norms, respectively, and
$\lambda>0$ is the ridge regularization coefficient.
Each weight $a_i$ scales the entire spectral fitting loss
of its corresponding sample, limiting the influence
of large fitting errors.

Each refitting procedure starts with an ordinary ridge
regression solution, followed by alternating sample-weight
estimation and weighted refitting until the stopping criterion
is met or the maximum number of iterations is reached.

\subsection{View-Specific Record Selection and Weight Estimation}
\label{app:tempra_details}

TempRA constructs long, short, and recent views from
the causal stream of matured feedback $\mathcal M_t$ to estimate
the reliability of the two evidence branches.
We index these views by $v\in\{L,S,R\}$, respectively.
Each record contains the AnalogE forecast $Y_i^A$ and RecentE
forecast $Y_i^R$ issued at the original forecast origin,
together with the subsequently observed outcome $Y_i$.
Reliability is evaluated using these issued forecasts;
historical predictions are not recomputed with updated modules.

\paragraph{Feedback Record Selection.}
Let $\mathcal S_t\subseteq\mathcal M_t$ denote the sparsely
selected mature records, with consecutive selected forecast
origins separated by at least $H$ time steps.
The record sets for the three views are
\begin{equation}
\mathcal V_t^L=\mathcal S_t,
\qquad
\mathcal V_t^S
=\operatorname{Last}_{n_S}(\mathcal S_t),
\qquad
\mathcal V_t^R
=\operatorname{Last}_{n_R}(\mathcal M_t),
\end{equation}
where $\operatorname{Last}_m$ returns the latest available
records, up to $m$, ordered by forecast origin.
The parameters $n_S$ and $n_R$ specify the maximum numbers
of records in the short and recent views, respectively.

The long view uses all sparsely selected records
to capture cumulative performance.
The short view uses the latest $n_S$ records from this set.
The recent view directly uses the latest $n_R$ records
from the dense feedback stream without the sparse selection
interval, capturing the most recent performance differences.

\paragraph{Long- and Short-View Weights.}
For branch $k\in\{A,R\}$, the forecast MSE on record $i$ is
\begin{equation}
\ell_i^k
=
\frac{1}{H}\left\|Y_i^k-Y_i\right\|_2^2.
\end{equation}
We normalize this loss by the error of a persistence forecast:
\begin{equation}
\ell_i^{\mathrm{pers}}
=
\frac{1}{H}
\left\|x_i^{\mathrm{last}}\mathbf 1_H-Y_i\right\|_2^2,
\qquad
\bar\ell_i^k
=
\frac{\ell_i^k}{\ell_i^{\mathrm{pers}}+\epsilon},
\end{equation}
where the persistence forecast uses the last context observation
$x_i^{\mathrm{last}}$ at every forecast step,
$\mathbf 1_H$ is a length-$H$ vector of ones, and
$\epsilon>0$ ensures numerical stability.

For view $v\in\{L,S\}$, we accumulate the normalized losses
over its record set and obtain the AnalogE weight through
exponential weighting:
\begin{equation}
L_{t,v}^{k}
=
\sum_{i\in\mathcal V_t^v}\bar\ell_i^k,
\qquad
w_t^v
=
\frac{\exp(-\eta L_{t,v}^{A})}
{\exp(-\eta L_{t,v}^{A})+\exp(-\eta L_{t,v}^{R})}.
\end{equation}
Here, $\eta>0$ controls the sensitivity of the weights
to differences in accumulated loss.
The corresponding RecentE weight is $1-w_t^v$.
When the record set is empty, the two branches receive
equal weights.

\paragraph{Recent-View Weight.}
The recent view accounts for both the magnitude and joint
behavior of the two branches' errors.
The historical forecast errors are
\begin{equation}
e_{i,h}^{A}=Y_{i,h}^{A}-Y_{i,h},
\qquad
e_{i,h}^{R}=Y_{i,h}^{R}-Y_{i,h},
\end{equation}
where $h$ denotes a forecast step.
We estimate the error second moments and cross-moment
using the recent feedback records:
\begin{equation}
\begin{aligned}
m_t^{AA}
&=
\frac{1}{N_t}
\sum_{i\in\mathcal V_t^R}
\sum_{h=1}^{H}(e_{i,h}^{A})^2,\\
m_t^{RR}
&=
\frac{1}{N_t}
\sum_{i\in\mathcal V_t^R}
\sum_{h=1}^{H}(e_{i,h}^{R})^2,\\
m_t^{AR}
&=
\frac{1}{N_t}
\sum_{i\in\mathcal V_t^R}
\sum_{h=1}^{H}e_{i,h}^{A}e_{i,h}^{R},
\end{aligned}
\end{equation}
where $N_t=|\mathcal V_t^R|H$ is the number of error terms
used for estimation.
These statistics are computed without mean centering
and therefore represent second moments and a cross-moment.

For an AnalogE weight $w$, the empirical combination error is
\begin{equation}
J_t(w)
=
w^2m_t^{AA}
+(1-w)^2m_t^{RR}
+2w(1-w)m_t^{AR}.
\end{equation}
The recent-view weight assigned to AnalogE minimizes this error:
\begin{equation}
w_t^R
=
\arg\min_{w\in[0,1]}J_t(w).
\end{equation}
When $m_t^{AA}+m_t^{RR}-2m_t^{AR}>0$, the solution is
\begin{equation}
w_t^R
=
\operatorname{clip}_{[0,1]}
\left(
\frac{m_t^{RR}-m_t^{AR}}
{m_t^{AA}+m_t^{RR}-2m_t^{AR}}
\right),
\end{equation}
where $\operatorname{clip}_{[0,1]}$ restricts the weight
to $[0,1]$.
Here, the superscript $R$ in $w_t^R$ denotes the recent view,
whereas $R$ in the forecast errors and moments denotes
the RecentE branch.
The corresponding RecentE weight is $1-w_t^R$.

When the recent view contains fewer than $n_{\min}$ mature
records, where $n_{\min}$ is the minimum required record count,
we skip moment estimation and assign equal weights
to the two branches.
If the denominator is zero, the combination error is
independent of the weight, and equal weighting is also
an optimal solution.

\paragraph{Multi-View Aggregation.}
The final AnalogE weight combines the three view-specific weights:
\begin{equation}
w_t
=
\lambda_L w_t^L
+
\lambda_S w_t^S
+
\lambda_R w_t^R,
\end{equation}
where the view-combination coefficients satisfy
\begin{equation}
\lambda_v\geq0,
\qquad
\sum_{v\in\{L,S,R\}}\lambda_v=1.
\end{equation}
The corresponding RecentE weight is $1-w_t$,
and the final prediction is
\begin{equation}
\hat Y_t
=
w_tY_t^A
+
(1-w_t)Y_t^R.
\end{equation}
\section{Experimental Setup and Implementation Details}
\label{app:experimental_details}

\subsection{Datasets and Data Usage}
\label{app:datasets}

We evaluate ECHO on time series datasets covering
electricity transformer measurements, exchange rates,
electricity consumption, solar power generation, weather,
and traffic flow.
We additionally evaluate on PEMS04 in the supplementary
cross-backbone experiments.

\paragraph{ETT.}
ETTh1, ETTh2, ETTm1, and ETTm2~\citep{zhou2021informer}
contain oil temperature and load measurements from two
electricity transformers.
ETTh1 and ETTh2 are sampled hourly, while ETTm1 and ETTm2
are sampled at 15-minute intervals.

\paragraph{Exchange Rate.}
Exchange Rate~\citep{lai2018modeling} contains daily exchange
rates of eight countries, providing a financial forecasting
setting with irregular fluctuations and changing temporal patterns.

\paragraph{Electricity and Solar-Energy.}
Electricity~\citep{lai2018modeling} records hourly electricity
consumption from 321 clients.
Solar-Energy~\citep{lai2018modeling} records solar power generation
from 137 photovoltaic plants at 10-minute intervals.
These datasets represent energy demand and renewable generation,
respectively.

\paragraph{Weather.}
Weather~\citep{wu2021autoformer} contains 21 meteorological
variables recorded at 10-minute intervals, including temperature,
humidity, and atmospheric pressure.
It covers variables with diverse physical meanings and
temporal dynamics.

\paragraph{PEMS.}
PEMS08 and PEMS04~\citep{song2020spatial} contain traffic flow
measurements from 170 and 307 sensors, respectively,
at 5-minute intervals.
They capture recurring traffic patterns and short-term
fluctuations in traffic flow.

\paragraph{Data Splits.}
We split each dataset chronologically into training, validation,
and test sets, using a ratio of 6:2:2 for the ETT datasets
and 7:1:2 for the remaining datasets.

\paragraph{Data Usage.}
We pretrain AnalogE using queries sampled from the Chronos
pretraining corpus~\citep{ansari2024chronos} and keep it frozen
on all downstream datasets.
For each downstream dataset, the training split is used to
construct a fixed retrieval memory of context--future pairs
and initialize the RecentE residual map.
The retrieval memory remains unchanged throughout validation
and testing.

The validation split provides causal online warm-up for RecentE
and TempRA.
The test split inherits their states at the end of validation
and is evaluated chronologically.
Online updates use only feedback whose complete target horizons
have been observed by the current forecast origin.
Validation and test feedback is neither added to the retrieval
memory nor used to update AnalogE.

\paragraph{Normalization and Evaluation Scale.}
Each variate is standardized using statistics estimated
from the training split only.
Query and candidate windows are then independently normalized
using their own context statistics, while the base forecast
uses the query statistics.
Context standard deviations are floored at $10^{-3}$,
and normalized values are clipped to $[-20,20]$.
Corrected forecasts are mapped back to the dataset-standardized
scale before computing MSE and MAE.
Macro averages are unweighted means of dataset-level metrics.

\subsection{ECHO Implementation and Training}
\label{app:echo_implementation}

\paragraph{Default Configuration.}
Unless otherwise specified, experiments use frozen TimesFM 2.5
with context length $L=512$, forecast horizon $H=64$, and
$K=20$ retrieved cases.
Forecast correction is performed independently for each variate.
AnalogE remains frozen during downstream evaluation, while
RecentE and TempRA use only causally available, fully matured feedback.

\paragraph{AnalogE Architecture and Retrieval.}
AnalogE uses separate candidate-set and relation encoders,
each with hidden dimension 32 and one four-head self-attention
layer.
The candidate-set encoder uses masked mean pooling to produce
a set summary, while the relation encoder supplies candidate
representations to the scoring head.
The learned score adjustment is bounded by $s_{\max}=0.5$,
with distance temperature $\tau=1$.
The gate combines the set summary with evidence statistics
through an MLP with hidden dimension 64, and its sigmoid
output is scaled by $g_{\max}=0.5$.
In implementation, the 19 explicit statistics are supplemented
with auxiliary local features.
The 64-step horizon is divided into eight contiguous groups.
Candidate weights and correction gates are computed separately
for each group, with network parameters shared across groups.
Group indices are omitted in the main-text formulation for clarity.
AnalogE has approximately 42K parameters, excluding the frozen
backbone and retrieval encoder.
Retrieval uses the 768-dimensional last-token representation
of the frozen \texttt{amazon/chronos-t5-base} encoder and
selects $K=20$ cases by squared Euclidean distance.
Downstream memories contain only complete training-split cases
whose futures are fully observed by the current forecast origin.

\paragraph{AnalogE Pretraining.}
For each backbone, we pretrain a separate AnalogE checkpoint
from scratch on 50,000 cross-domain examples from the Chronos
pretraining corpus and reuse it across all downstream datasets
without parameter updates.
Training uses AdamW with learning rate $10^{-4}$,
weight decay $10^{-4}$, batch size 32, and three epochs,
with cosine learning-rate scheduling and gradient-norm
clipping at 0.5.
The objective combines forecast MSE in the window-normalized
space with a gate-supervision MSE weighted by 0.1.
For each forecast group, the gate target is the least-squares
coefficient for correcting the base forecast toward the
historical reference, clipped to $[0,g_{\max}]$.
This target is used only during offline training.
The best checkpoint is selected by normalized forecast MSE
on the upstream validation set.

\paragraph{RecentE.}
The residual map is initialized on the target-domain training split.
The standard record-selection interval is $s=64$ forecast origins,
and refitting is triggered by $u=64$ newly selected, fully matured
feedback records, as defined in Appendix~\ref{app:recent_updates}.
Robust fitting uses groups of at most eight forecast steps,
a Huber threshold of $\delta=0.5$, and ridge regularization
$\lambda=10^{-3}$.
Each refitting procedure uses at most five reweighting iterations
with an early-stopping tolerance of $10^{-7}$.

\paragraph{TempRA.}
The short and recent views use up to 8 sparse records and
32 dense records, respectively.
The long and short views use exponential weighting with $\eta=1$.
The view-combination coefficients are selected on the validation
set and held fixed during testing.
Full record-selection and weight-estimation rules are given in
Appendix~\ref{app:tempra_details}.

\subsection{Long-Horizon Forecasting Protocol}
\label{app:long_horizon_protocol}

Long-horizon experiments use frozen TimesFM 2.5 with context
length $L=512$ and horizons $H\in\{96,192,336,720\}$.
We treat the model as a black-box forecaster.
Since the interface used in our experiments returns at most
128 steps per call, longer forecasts are generated recursively
by appending predicted values to the input and retaining
the latest 512 steps as context.

AnalogE follows a separate 64-step rollout without retraining.
At each step, it retrieves historical cases using its current
rollout context and corrects the resulting forecast.
The corrected segment is appended to its own context,
again retaining the latest 512 steps.
Only the required steps of the final segment are retained.
RecentE uses a residual map defined for the complete horizon $H$,
taking the original observed context and the complete base
forecast as inputs.

For a forecast issued at origin $t$, all rollout segments
use only observations and matured feedback available at $t$.
Internal rollout positions do not advance the feedback-availability
clock.
Historical forecasts become eligible for online updates only
after their complete $H$-step targets have been observed.

We refer to this long-horizon configuration as ECHO-Long.
Original and corrected forecasts are evaluated on the same
test queries and targets.
Complete results are reported in Appendix~\ref{app:long_horizon}.

\subsection{Baseline Configurations}
\label{app:baseline_configs}

All baselines follow the same data splits, evaluation queries,
normalization, and causal feedback availability as ECHO.
Below, we describe the adaptations made to each method.

\paragraph{BB-TS-RAG.}
Original TS-RAG integrates retrieved information with internal
backbone representations before the output projection.
To support black-box access, we move this integration to the
output side.
Retrieved futures are pooled using distance-based weights,
and a lightweight gating network uses the current context,
base forecast, and retrieval reference to produce a scalar
weight for their convex combination.
The retrieved cases, pretraining data, and training budget
are matched to AnalogE.

\paragraph{Naive RAF.}
We adapt the originally Chronos-based Naive RAF input
augmentation to frozen TimesFM, using the same retrieved
cases as AnalogE.
We retain its procedure of averaging retrieved segments,
normalizing and shifting the resulting segment for alignment,
and prepending it to the current context.
No trainable adapter is introduced.

\paragraph{ELF and ELF+$Y^{\mathrm{base}}$.}
ELF retains its original frequency-domain forecaster and
slow, fast, and merge weighters.
It uses a frequency-retention ratio of 0.9, Woodbury updates
with interval $M=200$, $L_2$ regularization $\lambda=20$,
a fast-weighting window of $B=5$, and five block updates
for warm-up.
The forecaster directly predicts future values, which are
dynamically combined with the frozen backbone forecast.
Evaluation follows ECHO's data splits, normalization,
test queries, and fully matured feedback constraints.
ELF+$Y^{\mathrm{base}}$ only adds the base forecast to
the forecaster inputs; all remaining components are unchanged.

\paragraph{Ada-Y and Ada-Y+$X$.}
Ada-Y retains its output adapter, which generates additive
corrections from the base forecast, with the optional router
disabled.
Ada-Y+$X$ changes only the input to the concatenation
of the current context and base forecast.
Its remaining architecture, initialization, and online update
settings are identical to Ada-Y.
Both variants retain a gradient update for each fully
matured feedback record.

\subsection{Experimental Environment and Runtime Measurement}
Experiments use one NVIDIA H200 GPU and 16 allocated CPU cores.
Accuracy evaluation uses a batch size of 32.
Runtime measurements use a batch size of 1, averaging
over 100 queries after 10 warm-up queries.

\section{Additional Experimental Results}
\label{app:additional_results}

This section provides complete forecasting results and additional
analyses of ECHO, covering cross-backbone performance, long-horizon
forecasting, evidence construction and aggregation, and robustness
to random seeds.

\subsection{Cross-Backbone Evaluation}
\label{app:cross_backbone}

We examine whether ECHO improves forecasting across different
frozen backbones and target datasets.
We evaluate TimesFM 2.5, Chronos-2, Chronos-Bolt, Moirai 2.0,
and Sundial, comparing each backbone's original forecasts
with their correction by ECHO.
Within each comparison, the backbone, data splits, test queries,
context length $L=512$, and forecasting horizon $H=64$ remain
unchanged.
ECHO accesses only the backbone's numerical inputs and forecast
outputs, and online updates use only fully observed feedback.

Table~\ref{tab:cross_backbone_full} reports complete MSE and MAE
results on the nine main datasets and the additional PEMS04 dataset.
ECHO reduces MSE across all evaluated backbone--dataset combinations.
On PEMS04, the relative MSE reductions range from 5.81\% to
32.74\% across the five backbones.
MAE also improves in many settings, although small increases
occur in some cases.
These results support the applicability of ECHO across different
backbones and domains, with particularly consistent improvements
in the primary MSE metric.

\begin{table*}[t]
\caption{
Complete cross-backbone forecasting results on the nine main datasets
and the additional PEMS04 dataset.
Ori denotes the frozen backbone, and +ECHO denotes its correction by ECHO.
Lower values are better.
The better result within each Ori/+ECHO pair is bolded,
including ties at the displayed precision.}
\label{tab:cross_backbone_full}
\centering
\setlength{\tabcolsep}{2.2pt}
\renewcommand{\arraystretch}{1.08}

\resizebox{\textwidth}{!}{%
\begin{tabular}{l*{20}{c}}
\toprule
\textbf{Dataset}
& \multicolumn{4}{c}{\textbf{TimesFM 2.5}}
& \multicolumn{4}{c}{\textbf{Chronos-2}}
& \multicolumn{4}{c}{\textbf{Chronos-Bolt}}
& \multicolumn{4}{c}{\textbf{Moirai 2.0}}
& \multicolumn{4}{c}{\textbf{Sundial}}
\\

\cmidrule(lr){2-5}
\cmidrule(lr){6-9}
\cmidrule(lr){10-13}
\cmidrule(lr){14-17}
\cmidrule(lr){18-21}

& \multicolumn{2}{c}{Ori} & \multicolumn{2}{c}{+ECHO}
& \multicolumn{2}{c}{Ori} & \multicolumn{2}{c}{+ECHO}
& \multicolumn{2}{c}{Ori} & \multicolumn{2}{c}{+ECHO}
& \multicolumn{2}{c}{Ori} & \multicolumn{2}{c}{+ECHO}
& \multicolumn{2}{c}{Ori} & \multicolumn{2}{c}{+ECHO}
\\

\cmidrule(lr){2-3}
\cmidrule(lr){4-5}
\cmidrule(lr){6-7}
\cmidrule(lr){8-9}
\cmidrule(lr){10-11}
\cmidrule(lr){12-13}
\cmidrule(lr){14-15}
\cmidrule(lr){16-17}
\cmidrule(lr){18-19}
\cmidrule(lr){20-21}

& MSE & MAE & MSE & MAE
& MSE & MAE & MSE & MAE
& MSE & MAE & MSE & MAE
& MSE & MAE & MSE & MAE
& MSE & MAE & MSE & MAE
\\
\midrule

ETTh1
& .355 & \textbf{.369} & \textbf{.346} & .371
& .377 & \textbf{.366} & \textbf{.348} & \textbf{.366}
& .362 & \textbf{.365} & \textbf{.343} & \textbf{.365}
& .367 & .367 & \textbf{.340} & \textbf{.365}
& .337 & \textbf{.369} & \textbf{.336} & .370
\\

ETTh2
& .257 & .315 & \textbf{.242} & \textbf{.310}
& .266 & \textbf{.301} & \textbf{.246} & .302
& .252 & \textbf{.299} & \textbf{.240} & .302
& .266 & .306 & \textbf{.241} & \textbf{.305}
& .242 & .308 & \textbf{.233} & \textbf{.306}
\\

ETTm1
& .280 & .316 & \textbf{.263} & \textbf{.314}
& .309 & .310 & \textbf{.274} & \textbf{.303}
& .311 & .318 & \textbf{.271} & \textbf{.312}
& .322 & .321 & \textbf{.273} & \textbf{.314}
& .268 & \textbf{.309} & \textbf{.257} & .310
\\

ETTm2
& .148 & .232 & \textbf{.136} & \textbf{.225}
& .149 & .219 & \textbf{.138} & \textbf{.218}
& .149 & .224 & \textbf{.139} & \textbf{.222}
& .155 & .228 & \textbf{.138} & \textbf{.223}
& .149 & .234 & \textbf{.135} & \textbf{.224}
\\

Exchange
& .059 & .168 & \textbf{.056} & \textbf{.164}
& .060 & .167 & \textbf{.058} & \textbf{.166}
& .067 & .178 & \textbf{.064} & \textbf{.174}
& .060 & .169 & \textbf{.058} & \textbf{.167}
& .082 & .196 & \textbf{.067} & \textbf{.176}
\\

Weather
& .135 & .171 & \textbf{.124} & \textbf{.164}
& .143 & .162 & \textbf{.128} & \textbf{.160}
& .152 & .182 & \textbf{.134} & \textbf{.174}
& .148 & .172 & \textbf{.127} & \textbf{.164}
& .136 & .179 & \textbf{.123} & \textbf{.166}
\\

Electricity
& .120 & .212 & \textbf{.116} & \textbf{.209}
& .116 & \textbf{.200} & \textbf{.113} & \textbf{.200}
& .113 & \textbf{.200} & \textbf{.111} & \textbf{.200}
& .117 & .205 & \textbf{.114} & \textbf{.203}
& .107 & .202 & \textbf{.106} & \textbf{.201}
\\

PEMS04
& .112 & .212 & \textbf{.104} & \textbf{.207}
& .219 & .285 & \textbf{.167} & \textbf{.259}
& .255 & .317 & \textbf{.171} & \textbf{.267}
& .127 & .223 & \textbf{.112} & \textbf{.215}
& .112 & .208 & \textbf{.105} & \textbf{.205}
\\

PEMS08
& .097 & .196 & \textbf{.089} & \textbf{.191}
& .232 & .295 & \textbf{.179} & \textbf{.268}
& .289 & .338 & \textbf{.190} & \textbf{.278}
& .112 & .212 & \textbf{.098} & \textbf{.203}
& .094 & .193 & \textbf{.088} & \textbf{.189}
\\

Solar
& .208 & \textbf{.217} & \textbf{.180} & \textbf{.217}
& .147 & \textbf{.162} & \textbf{.145} & .172
& .469 & .337 & \textbf{.359} & \textbf{.297}
& .181 & \textbf{.197} & \textbf{.159} & .202
& .210 & .225 & \textbf{.189} & \textbf{.224}
\\

\bottomrule
\end{tabular}%
}
\end{table*}

\subsection{Long-Horizon Forecasting}
\label{app:long_horizon}

We examine whether ECHO's forecasting improvements extend
to longer prediction horizons.
Using frozen TimesFM 2.5 with black-box Rolling-128 forecasting,
we compare the original forecasts with those corrected by
ECHO-Long at $H\in\{96,192,336,720\}$.
The context length remains fixed at $L=512$.
For each horizon, both configurations use the same test queries
and target values, while online adaptation uses only feedback
whose complete target horizons have been observed.

Table~\ref{tab:long_horizon_full} reports the results on nine datasets.
ECHO-Long reduces macro-average MSE by 6.91\%, 7.11\%, 8.10\%,
and 8.50\% at the four horizons, respectively.
MSE improves in 34 of the 36 dataset--horizon combinations.

\begin{table*}[t]
\caption{
Complete long-horizon forecasting results of TimesFM 2.5 on nine
datasets using black-box forecasting.
The context length is $L=512$, with forecasting horizons
$H\in\{96,192,336,720\}$.
Ori denotes the original TimesFM 2.5 forecasts, and +ECHO denotes
their correction using ECHO-Long.
MSE and MAE are reported, with lower values indicating better performance.
The better result within each Ori/+ECHO pair is bolded.
The last row reports the relative change in the unweighted macro average
over the nine datasets.
Aggregates and relative changes are computed from unrounded values.
}
\label{tab:long_horizon_full}
\centering

\setlength{\tabcolsep}{2.2pt}
\renewcommand{\arraystretch}{1.08}

\resizebox{\textwidth}{!}{%
\begin{tabular}{l*{16}{c}}
\toprule

\textbf{Dataset}
& \multicolumn{4}{c}{\textbf{$H=96$}}
& \multicolumn{4}{c}{\textbf{$H=192$}}
& \multicolumn{4}{c}{\textbf{$H=336$}}
& \multicolumn{4}{c}{\textbf{$H=720$}}
\\

\cmidrule(lr){2-5}
\cmidrule(lr){6-9}
\cmidrule(lr){10-13}
\cmidrule(lr){14-17}

& \multicolumn{2}{c}{Ori}
& \multicolumn{2}{c}{+ECHO}
& \multicolumn{2}{c}{Ori}
& \multicolumn{2}{c}{+ECHO}
& \multicolumn{2}{c}{Ori}
& \multicolumn{2}{c}{+ECHO}
& \multicolumn{2}{c}{Ori}
& \multicolumn{2}{c}{+ECHO}
\\

\cmidrule(lr){2-3}
\cmidrule(lr){4-5}
\cmidrule(lr){6-7}
\cmidrule(lr){8-9}
\cmidrule(lr){10-11}
\cmidrule(lr){12-13}
\cmidrule(lr){14-15}
\cmidrule(lr){16-17}

& MSE & MAE & MSE & MAE
& MSE & MAE & MSE & MAE
& MSE & MAE & MSE & MAE
& MSE & MAE & MSE & MAE
\\

\midrule

ETTh1
& .3753 & .3826 & \textbf{.3637} & \textbf{.3825}
& .4149 & .4071 & \textbf{.3939} & \textbf{.4035}
& .4344 & .4183 & \textbf{.4103} & \textbf{.4140}
& .4191 & \textbf{.4269} & \textbf{.4062} & .4313
\\

ETTh2
& .2989 & .3429 & \textbf{.2763} & \textbf{.3339}
& .3681 & .3876 & \textbf{.3408} & \textbf{.3779}
& .3907 & .4101 & \textbf{.3691} & \textbf{.4018}
& .4043 & .4297 & \textbf{.3924} & \textbf{.4261}
\\

ETTm1
& .3011 & .3299 & \textbf{.2842} & \textbf{.3298}
& .3457 & \textbf{.3596} & \textbf{.3271} & .3602
& .3773 & \textbf{.3826} & \textbf{.3662} & .3899
& .4405 & \textbf{.4185} & \textbf{.4316} & .4323
\\

ETTm2
& .1741 & .2519 & \textbf{.1607} & \textbf{.2437}
& .2398 & .2979 & \textbf{.2187} & \textbf{.2855}
& .3018 & .3388 & \textbf{.2780} & \textbf{.3264}
& .4039 & .3989 & \textbf{.3682} & \textbf{.3846}
\\

Exchange
& .0914 & .2103 & \textbf{.0854} & \textbf{.2048}
& .1943 & .3144 & \textbf{.1812} & \textbf{.3028}
& .3597 & .4363 & \textbf{.3258} & \textbf{.4132}
& 1.0458 & .7496 & \textbf{.9443} & \textbf{.7445}
\\

Weather
& .1590 & .1990 & \textbf{.1437} & \textbf{.1910}
& .2094 & .2465 & \textbf{.1889} & \textbf{.2379}
& .2748 & .2915 & \textbf{.2465} & \textbf{.2830}
& .3620 & .3477 & \textbf{.3322} & \textbf{.3431}
\\

Electricity
& .1308 & \textbf{.2217} & \textbf{.1272} & .2220
& .1478 & \textbf{.2379} & \textbf{.1462} & .2430
& \textbf{.1655} & \textbf{.2560} & .1657 & .2647
& \textbf{.2053} & \textbf{.2911} & .2090 & .3049
\\

PEMS08
& .1091 & .2040 & \textbf{.0984} & \textbf{.1983}
& .1432 & .2244 & \textbf{.1256} & \textbf{.2157}
& .1772 & .2465 & \textbf{.1563} & \textbf{.2350}
& .2492 & .3082 & \textbf{.1991} & \textbf{.2763}
\\

Solar
& .2217 & \textbf{.2255} & \textbf{.1932} & .2384
& .2552 & \textbf{.2493} & \textbf{.2311} & .2696
& .2915 & \textbf{.2754} & \textbf{.2303} & .2865
& .3150 & \textbf{.2977} & \textbf{.2352} & .3020
\\

\midrule

\rowcolor{gray!8}
\textbf{Macro Avg.}
& .2068 & .2631 & \textbf{.1925} & \textbf{.2605}
& .2576 & .3027 & \textbf{.2393} & \textbf{.2996}
& .3081 & .3395 & \textbf{.2831} & \textbf{.3350}
& .4272 & .4076 & \textbf{.3909} & \textbf{.4050}
\\

\rowcolor{gray!14}
\textbf{Avg. Change}
& -- & --
& \imp{-6.91\%} & \imp{-0.99\%}
& -- & --
& \imp{-7.11\%} & \imp{-1.05\%}
& -- & --
& \imp{-8.10\%} & \imp{-1.33\%}
& -- & --
& \imp{-8.50\%} & \imp{-0.63\%}
\\

\bottomrule
\end{tabular}%
}
\end{table*}

\subsection{Complementarity of Historical and Recent Evidence}
\label{app:complementarity}

We examine whether jointly using historical analogues and
recent forecast errors provides more consistent corrections
than relying on either evidence source alone.
Using frozen TimesFM 2.5, we compare four configurations:
the uncorrected backbone, AnalogE-only, RecentE-only, and full ECHO.
AnalogE-only corrects the base forecast using historical analogues,
whereas RecentE-only applies the learned residual correction.
Full ECHO combines the two corrected forecasts through TempRA.
The standalone branches retain their respective configurations
from full ECHO, and all variants use the same base forecasts
and test queries.

Table~\ref{tab:echo_ablation} provides the complete results
underlying Figure~\ref{fig:evidence_radar}.
The stronger standalone branch varies across datasets.
AnalogE-only achieves lower MSE than RecentE-only on ETTh1,
ETTh2, ETTm1, Exchange, and Solar, while RecentE-only performs
better on ETTm2, Weather, Electricity, and PEMS08.
Full ECHO achieves the lowest MSE on seven of the nine datasets
and improves upon the backbone on all nine.
RecentE-only remains better on ETTm2 and Weather.
These findings support the complementarity of the two sources:
their joint use yields more consistent improvements, although
aggregation does not always outperform the stronger individual
branch.

 \begin{table*}[t]
  \caption{
  Ablation study of ECHO on TimesFM 2.5 across nine datasets with
  context length 512 and forecasting horizon 64.
  AnalogE Only and RecentE Only retain Historical Analog Evidence and
  Adaptive Recent Evidence, respectively.
  The corresponding radar visualization uses the same MSE results, with
  ETTh1/ETTh2 and ETTm1/ETTm2 averaged as ETTh Avg.\ and ETTm Avg.
  Each entry reports MSE/MAE; lower values are better.
  Best and second-best results are shown in \textbf{bold} and
  \underline{underlined}, respectively, according to unrounded values.
  }
  \label{tab:echo_ablation}
  \centering
  \small
  \setlength{\tabcolsep}{2.0pt}
  \renewcommand{\arraystretch}{1.08}

  \resizebox{\textwidth}{!}{%
  \begin{tabular}{@{}l*{18}{c}@{}}
  \toprule
  & \multicolumn{2}{c}{ETTh1}
  & \multicolumn{2}{c}{ETTh2}
  & \multicolumn{2}{c}{ETTm1}
  & \multicolumn{2}{c}{ETTm2}
  & \multicolumn{2}{c}{Exchange}
  & \multicolumn{2}{c}{Weather}
  & \multicolumn{2}{c}{Electricity}
  & \multicolumn{2}{c}{PEMS08}
  & \multicolumn{2}{c}{Solar}

  \\

  \cmidrule(lr){2-3}
  \cmidrule(lr){4-5}
  \cmidrule(lr){6-7}
  \cmidrule(lr){8-9}
  \cmidrule(lr){10-11}
  \cmidrule(lr){12-13}
  \cmidrule(lr){14-15}
  \cmidrule(lr){16-17}
  \cmidrule(lr){18-19}

  \textbf{Method}
  & MSE & MAE
  & MSE & MAE
  & MSE & MAE
  & MSE & MAE
  & MSE & MAE
  & MSE & MAE
  & MSE & MAE
  & MSE & MAE
  & MSE & MAE

  \\
  \midrule

  Backbone
  & .355 & \underline{.369}
  & .257 & .315
  & .280 & \underline{.316}
  & .148 & .232
  & .059 & .168
  & .135 & .171
  & .120 & .212
  & .097 & .196
  & .208 & \textbf{.217}

  \\

  AnalogE Only
  & \underline{.347} & \textbf{.369}
  & \underline{.246} & \underline{.313}
  & \underline{.268} & .318
  & .143 & .231
  & \underline{.056} & \underline{.165}
  & .131 & .172
  & .119 & \underline{.212}
  & .103 & .209
  & \underline{.181} & \underline{.217}

  \\

  RecentE Only
  & .364 & .389
  & .249 & .319
  & .270 & .319
  & \textbf{.135} & \underline{.225}
  & .070 & .188
  & \textbf{.122} & \underline{.165}
  & \underline{.118} & .214
  & \underline{.089} & \underline{.193}
  & .186 & .246

  \\

  \midrule

  \rowcolor{gray!10}
  \textbf{ECHO}
  & \textbf{.346} & .371
  & \textbf{.242} & \textbf{.310}
  & \textbf{.263} & \textbf{.314}
  & \underline{.136} & \textbf{.225}
  & \textbf{.056} & \textbf{.164}
  & \underline{.124} & \textbf{.164}
  & \textbf{.116} & \textbf{.209}
  & \textbf{.089} & \textbf{.191}
  & \textbf{.180} & .217

  \\

  \bottomrule
  \end{tabular}%
  }
  \end{table*}

\subsection{Construction of Historical Analog Evidence}
\label{app:analog_construction}

We examine the effectiveness of AnalogE in using retrieved
historical cases for forecast correction.
We compare standalone AnalogE with BB-TS-RAG and Naive RAF
on frozen TimesFM 2.5.
BB-TS-RAG is a black-box-compatible adaptation of TS-RAG,
whose original formulation integrates retrieved representations
with internal TSFM representations.
Naive RAF keeps the backbone frozen.
RecentE and TempRA are excluded so that the comparison focuses
on historical evidence construction.

Table~\ref{tab:analog_evidence} reports the complete results
on nine datasets.
AnalogE achieves the lowest MSE on seven datasets.
Naive RAF performs better on Weather, while the uncorrected
backbone remains best on PEMS08.
These results support AnalogE's effectiveness in constructing
corrections from historical analogues.
The PEMS08 results also show that historical evidence alone
can be insufficient, consistent with the benefit of incorporating
recent forecast-error evidence.

\begin{table*}[t]
  \caption{
  Comparison of historical evidence integration methods on a frozen
  TimesFM 2.5 backbone.
  Naive RAF uses a frozen backbone.
  ECHO AnalogE denotes our Historical Analog Evidence module.
  Each result is reported as MSE/MAE; lower values are better.
  Best results are bolded according to unrounded values.
  }
  \label{tab:analog_evidence}
  \centering
  \small
  \setlength{\tabcolsep}{3.0pt}
  \renewcommand{\arraystretch}{1.08}

  \resizebox{\textwidth}{!}{%
  \begin{tabular}{l*{9}{c}}
  \toprule
  \textbf{Method}
  & \textbf{ETTh1}
  & \textbf{ETTh2}
  & \textbf{ETTm1}
  & \textbf{ETTm2}
  & \textbf{Exchange}
  & \textbf{Weather}
  & \textbf{Electricity}
  & \textbf{PEMS08}
  & \textbf{Solar}
  \\
  \midrule

  TimesFM 2.5
  & .355 / .369
  & .257 / .315
  & .280 / .316
  & .148 / .232
  & .059 / .168
  & .135 / .171
  & .120 / .212
  & \textbf{.097} / \textbf{.196}
  & .208 / .217

  \\

  \midrule

  BB-TS-RAG
  & .351 / .369
  & .247 / .310
  & .272 / \textbf{.315}
  & .152 / .235
  & .074 / .183
  & .139 / .180
  & .127 / .214
  & .101 / .202
  & .196 / .221

  \\

  Naive RAF
  & .350 / .372
  & .248 / \textbf{.306}
  & .280 / .319
  & .146 / \textbf{.230}
  & .058 / .166
  & \textbf{.130} / \textbf{.167}
  & .127 / .221
  & .137 / .238
  & .202 / \textbf{.208}

  \\

  \midrule

  \rowcolor{gray!10}
  ECHO AnalogE
  & \textbf{.347} / \textbf{.369}
  & \textbf{.246} / .313
  & \textbf{.268} / .318
  & \textbf{.143} / .231
  & \textbf{.056} / \textbf{.165}
  & .131 / .172
  & \textbf{.119} / \textbf{.212}
  & .103 / .209
  & \textbf{.181} / .217

  \\

  \bottomrule
  \end{tabular}%
  }
  \end{table*}

\subsection{Construction of Adaptive Recent Evidence}
\label{app:recent_construction}

We compare RecentE with ELF and Ada-Y, including variants
with matched predictor inputs.
ELF retains its original forecaster and ensemble weighting;
ELF+$Y^{\mathrm{base}}$ additionally provides the base forecast
to its forecaster.
Ada-Y+$X$ additionally provides the context to Ada-Y.
Each input-augmented variant otherwise retains its baseline settings.
RecentE uses both inputs to estimate residual corrections.
All methods use frozen TimesFM 2.5 and are evaluated without
AnalogE or TempRA.

Table~\ref{tab:recent_architecture} shows that RecentE achieves
the lowest MSE on ETTm1, ETTm2, and Weather, including against
alternatives with matched predictor inputs.
ELF variants achieve the lowest MSE on ETTh1 and ETTh2,
while Ada-Y+$X$ performs best on Exchange.
These results support RecentE's design beyond input access alone,
although its advantage varies across datasets and metrics.

\begin{table*}[t]
\caption{
MSE/MAE comparison of recent-evidence methods on six datasets
with TimesFM 2.5.
$X$ and $Y^{\mathrm{base}}$ indicate predictor inputs.
ELF retains its original forecaster and ensemble weighting.
All methods are evaluated without AnalogE or TempRA.
Lower is better.
}
\label{tab:recent_architecture}
\centering
\small
\setlength{\tabcolsep}{3.0pt}
\renewcommand{\arraystretch}{1.08}

\begin{tabular}{llcc*{6}{c}}
\toprule
\textbf{Family}
& \textbf{Variant}
& $X$
& $Y^{\mathrm{base}}$
& \textbf{ETTh1}
& \textbf{ETTh2}
& \textbf{ETTm1}
& \textbf{ETTm2}
& \textbf{Exchange}
& \textbf{Weather}
\\
\midrule

ELF
& --
& \checkmark
& --
& .342/.367
& .247/.314
& .276/.315
& .146/.232
& .067/.183
& .133/.171
\\

ELF
& +$Y^{\mathrm{base}}$
& \checkmark
& \checkmark
& .342/.367
& .250/.316
& .276/.316
& .147/.232
& .067/.184
& .133/.171
\\
\midrule

Ada-Y
& --
& --
& \checkmark
& .357/.370
& .257/.319
& .279/.316
& .149/.233
& .059/.171
& .134/.170
\\

Ada-Y
& +$X$
& \checkmark
& \checkmark
& .357/.371
& .258/.319
& .280/.317
& .148/.232
& .057/.167
& .133/.171
\\
\midrule

\rowcolor{gray!10}
RecentE
& --
& \checkmark
& \checkmark
& .364/.389
& .249/.319
& .270/.319
& .135/.225
& .070/.188
& .122/.165
\\
\bottomrule
\end{tabular}
\end{table*}

\subsection{Dynamic Aggregation versus Fixed Fusion}
\label{app:fixed_fusion}

We examine whether dynamically adjusting the contributions
of AnalogE and RecentE improves upon fixed fusion.
Using TimesFM 2.5 and Chronos-2, we compare TempRA with a fixed
fusion baseline whose weights are selected on the validation
split and held constant during testing.
The two methods combine exactly the same AnalogE and RecentE
forecasts on the same test queries.
Thus, the comparison isolates the aggregation rule:
fixed fusion retains validation-selected weights, whereas
TempRA updates the relative contributions using causally
available forecasting feedback.

Table~\ref{tab:fixed_fusion} shows that TempRA reduces
macro-average MSE by 0.81\% and 0.84\% for TimesFM 2.5 and
Chronos-2, respectively.
The corresponding MAE reductions are 1.48\% and 2.78\%.
The largest improvements occur on Exchange.
However, fixed fusion achieves lower MSE in some settings,
including ETTm2 and Weather for both backbones.
These results support a modest average benefit from dynamic
aggregation, while showing that fixed weights remain
competitive on individual datasets.

\begin{table*}[t]
\caption{
Comparison of TempRA with validation-selected fixed fusion.
Macro reports the average over the six datasets.
Relative Change reports the change in macro-average MSE/MAE
from fixed fusion to ECHO.
Lower values are better; the best results within each backbone
are bolded, including ties at the displayed precision.
Aggregates and relative changes are computed from unrounded values.
}
\label{tab:fixed_fusion}
\centering
\small
\setlength{\tabcolsep}{6pt}
\renewcommand{\arraystretch}{1.12}

\begin{tabular}{l*{8}{c}}
\toprule
\textbf{Dataset}
& \multicolumn{4}{c}{\textbf{TimesFM 2.5}}
& \multicolumn{4}{c}{\textbf{Chronos-2}}
\\

\cmidrule(lr){2-5}
\cmidrule(lr){6-9}

& \multicolumn{2}{c}{Fixed Fusion}
& \multicolumn{2}{c}{ECHO}
& \multicolumn{2}{c}{Fixed Fusion}
& \multicolumn{2}{c}{ECHO}
\\

\cmidrule(lr){2-3}
\cmidrule(lr){4-5}
\cmidrule(lr){6-7}
\cmidrule(lr){8-9}

& MSE & MAE & MSE & MAE
& MSE & MAE & MSE & MAE
\\
\midrule

ETTh1
& .352 & .380 & \textbf{.346} & \textbf{.371}
& .355 & .380 & \textbf{.348} & \textbf{.366}
\\

ETTh2
& .243 & .312 & \textbf{.242} & \textbf{.310}
& \textbf{.241} & .305 & .246 & \textbf{.302}
\\

ETTm1
& .264 & .315 & \textbf{.263} & \textbf{.314}
& \textbf{.266} & .311 & .274 & \textbf{.303}
\\

ETTm2
& \textbf{.133} & \textbf{.223} & .136 & .225
& \textbf{.135} & \textbf{.218} & .138 & \textbf{.218}
\\

Exchange
& .064 & .179 & \textbf{.056} & \textbf{.164}
& .084 & .183 & \textbf{.058} & \textbf{.166}
\\

Weather
& \textbf{.121} & \textbf{.163} & .124 & .164
& \textbf{.122} & .161 & .128 & \textbf{.160}
\\

\midrule
\textbf{Macro}
& .196 & .262 & \textbf{.195} & \textbf{.258}
& .200 & .260 & \textbf{.199} & \textbf{.253}
\\

\rowcolor{gray!12}
\textbf{Relative Change}
& -- & -- & \textbf{-0.81\%} & \textbf{-1.48\%}
& -- & -- & \textbf{-0.84\%} & \textbf{-2.78\%}
\\

\bottomrule
\end{tabular}
\end{table*}

\subsection{Role of Temporal Views in TempRA}
\label{app:temporal_views}

We examine whether combining multiple temporal views improves
the reliability estimates used by TempRA.
The long, short, and recent views use the same historical
feedback stream but differ in temporal coverage and feedback
density, with the recent view using dense feedback.
We compare single-view configurations, which retain only
one view; leave-one-out configurations, which omit one view;
and full TempRA, which uses all three.
The AnalogE and RecentE forecasts remain fixed, so that
the comparison focuses on the temporal information used
to determine aggregation weights.

\begin{figure*}[t]
    \centering
    \includegraphics[width=\textwidth]
    {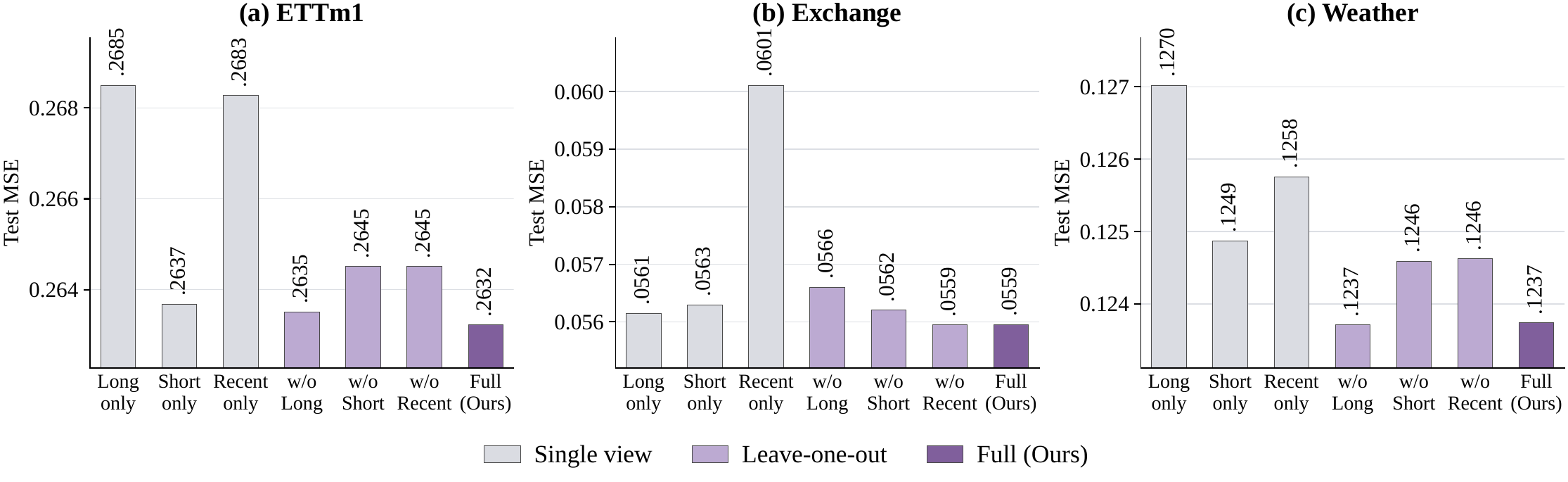}
    \caption{
    Ablation of temporal views in TempRA.
    Panels (a)--(c) report test MSE on ETTm1, Exchange, and Weather,
    respectively, comparing single-view, leave-one-out, and full TempRA
    configurations.
    Long, Short, and Recent denote the long, short, and recent views,
    with Recent using dense feedback.
    Each panel uses a different vertical axis range; lower MSE is better.
    }
    \label{fig:temporal_evidence_analysis}
\end{figure*}

Figure~\ref{fig:temporal_evidence_analysis} presents the results
on ETTm1, Exchange, and Weather.
The relative advantages of individual views vary across
datasets, and no single view consistently performs best.
Full TempRA achieves the best or near-best results across
the evaluated datasets.
These findings support using multiple temporal views to
combine reliability information from different portions
of the forecasting history.

\subsection{Robustness to Random Seeds}
\label{app:random_seeds}

We examine whether ECHO's improvements are sensitive to
the random seed used for AnalogE pretraining.
We evaluate three seeds, 42, 2024, and 2025, across five
backbones and six datasets: ETTh1, ETTh2, ETTm1, ETTm2,
Exchange, and Weather.
For each backbone and seed, we independently pretrain AnalogE
from scratch on 50,000 cross-domain examples for three epochs
and freeze the resulting checkpoint during downstream evaluation.
The data splits, fixed training-set retrieval memory,
retrieval size $K=20$, and RecentE and TempRA settings
remain unchanged.
Each run follows the same chronological validation warm-up
and test evaluation procedure.

Tables~\ref{tab:multiseed_mse} and~\ref{tab:multiseed_mae}
report the mean and sample standard deviation over the three
seeds for each dataset--backbone combination.
MSE standard deviations remain below 1.4\% of the corresponding
means, except for Sundial on Exchange.
MAE standard deviations remain below 1.1\% of the corresponding means.
ECHO also reduces MSE relative to the frozen backbone in all
90 dataset--backbone--seed configurations.
These results indicate low variability and consistent MSE
improvements across the evaluated seeds.

\begin{table*}[t]
\centering
\caption{
MSE of ECHO on six datasets across five forecasting backbones.
Each entry reports the mean $\pm$ sample standard deviation
over three seeds (42, 2024, and 2025), with AnalogE independently
pretrained for each seed. Lower values are better.
}
\label{tab:multiseed_mse}
\setlength{\tabcolsep}{4pt}
\renewcommand{\arraystretch}{1.15}
\resizebox{\textwidth}{!}{%
\begin{tabular}{lccccc}
\toprule
\textbf{Dataset}
& \textbf{TimesFM 2.5}
& \textbf{Chronos-2}
& \textbf{Chronos-Bolt}
& \textbf{Moirai 2.0}
& \textbf{Sundial} \\
\midrule
ETTh1
& $0.34601 \pm 0.00080$
& $0.34853 \pm 0.00040$
& $0.34358 \pm 0.00092$
& $0.33998 \pm 0.00108$
& $0.33314 \pm 0.00225$ \\
ETTh2
& $0.24253 \pm 0.00085$
& $0.24634 \pm 0.00084$
& $0.23981 \pm 0.00045$
& $0.24126 \pm 0.00053$
& $0.23371 \pm 0.00055$ \\
ETTm1
& $0.26401 \pm 0.00076$
& $0.27446 \pm 0.00074$
& $0.27301 \pm 0.00356$
& $0.27541 \pm 0.00181$
& $0.25828 \pm 0.00152$ \\
ETTm2
& $0.13644 \pm 0.00034$
& $0.13789 \pm 0.00029$
& $0.13941 \pm 0.00034$
& $0.13801 \pm 0.00015$
& $0.13519 \pm 0.00031$ \\
Exchange
& $0.05627 \pm 0.00054$
& $0.05847 \pm 0.00012$
& $0.06306 \pm 0.00047$
& $0.05785 \pm 0.00024$
& $0.06588 \pm 0.00126$ \\
Weather
& $0.12380 \pm 0.00028$
& $0.12841 \pm 0.00035$
& $0.13400 \pm 0.00046$
& $0.12779 \pm 0.00080$
& $0.12290 \pm 0.00015$ \\
\bottomrule
\end{tabular}%
}
\end{table*}

\begin{table*}[t]
\centering
\caption{
MAE of ECHO on six datasets across five forecasting backbones.
Each entry reports the mean $\pm$ sample standard deviation
over three seeds (42, 2024, and 2025), with AnalogE independently
pretrained for each seed. Lower values are better.
}
\label{tab:multiseed_mae}
\setlength{\tabcolsep}{4pt}
\renewcommand{\arraystretch}{1.15}
\resizebox{\textwidth}{!}{%
\begin{tabular}{lccccc}
\toprule
\textbf{Dataset}
& \textbf{TimesFM 2.5}
& \textbf{Chronos-2}
& \textbf{Chronos-Bolt}
& \textbf{Moirai 2.0}
& \textbf{Sundial} \\
\midrule
ETTh1
& $0.37077 \pm 0.00052$
& $0.36652 \pm 0.00013$
& $0.36462 \pm 0.00006$
& $0.36536 \pm 0.00028$
& $0.36795 \pm 0.00146$ \\
ETTh2
& $0.30948 \pm 0.00132$
& $0.30042 \pm 0.00115$
& $0.30017 \pm 0.00149$
& $0.30414 \pm 0.00113$
& $0.30505 \pm 0.00065$ \\
ETTm1
& $0.31487 \pm 0.00119$
& $0.30361 \pm 0.00054$
& $0.31177 \pm 0.00046$
& $0.31358 \pm 0.00003$
& $0.30899 \pm 0.00109$ \\
ETTm2
& $0.22421 \pm 0.00049$
& $0.21674 \pm 0.00091$
& $0.22051 \pm 0.00120$
& $0.22241 \pm 0.00024$
& $0.22374 \pm 0.00054$ \\
Exchange
& $0.16422 \pm 0.00075$
& $0.16562 \pm 0.00021$
& $0.17289 \pm 0.00067$
& $0.16712 \pm 0.00050$
& $0.17544 \pm 0.00183$ \\
Weather
& $0.16390 \pm 0.00028$
& $0.15913 \pm 0.00085$
& $0.17201 \pm 0.00145$
& $0.16418 \pm 0.00026$
& $0.16605 \pm 0.00018$ \\
\bottomrule
\end{tabular}%
}
\end{table*}

\end{document}